\documentclass[11pt]{article}

\usepackage[final]{acl}

\usepackage{times}
\usepackage{latexsym}
\usepackage{adjustbox}
\usepackage{booktabs}

\usepackage{array}
\usepackage{multirow}
\usepackage{xcolor}
\usepackage{colortbl}

\usepackage{vcell}
\usepackage{tabularx}
\usepackage{color}
\usepackage{tabularray}

\usepackage[T1]{fontenc}

\usepackage[utf8]{inputenc}

\usepackage{microtype}

\usepackage{inconsolata}

\usepackage{graphicx}

\usepackage[normalem]{ulem}

\title{Fine PT-PT Web: A High-Quality 41 Billion Tokens Data Collection of the\\
European Portuguese Web}

\author{
  \textbf{Gonçalo Vinagre\textsuperscript{1,2}},
  \textbf{Rui Pedro Guerra\textsuperscript{1,3}},
  \textbf{Pedro Gomes\textsuperscript{3}},
  \textbf{Miguel Moura Ramos\textsuperscript{4,5}},
\\
  \textbf{Duarte Miguel Alves\textsuperscript{4,5}},
  \textbf{Afonso Simplício\textsuperscript{1,2}},
  \textbf{Diogo Tavares\textsuperscript{1,2}},
  \textbf{David Semedo\textsuperscript{1,2}},
\\
  \textbf{Daniel Gomes\textsuperscript{3}},
  \textbf{João Magalhães\textsuperscript{1,2}}
\\
\\
 \textsuperscript{1}NOVA School of Science and Technology,
 \textsuperscript{2}NOVA LINCS, 
 \textsuperscript{3}Fundação para a Ciência e Tecnologia,
 \\
 \textsuperscript{4}Instituto Superior Técnico, Universidade de Lisboa, 
 \textsuperscript{5}Instituto de Telecomunicações, 
\\
  \small{
    \textbf{Correspondence:} \href{mailto:gv.martins@fct.unl.pt}{gv.martins@fct.unl.pt}
  }
}

\begin{document}
\maketitle
\begin{abstract}

Curating Web corpora for regional language variants like European Portuguese (PT-PT) is heavily bottlenecked by dialectal overlap (mainly with PT-BR) and data processing scale. This paper presents an efficient pipeline to curate a production-ready PT-PT corpus from the Portuguese Web, spanning 411 TB of raw data from Arquivo.pt. We introduce a novel post-scraping block that removes boilerplate and line duplicates prior to filtering. This early-stage intervention increases final document yield by 19.04\% by rescuing valid text that standard heuristic filters prematurely discard. Integrated with rigorous language identification, weighted fuzzy deduplication, and neural quality classification, our pipeline offers a scalable framework and a clean, representative corpus optimized for LLM pre-training.\footnote{For the pipeline code, refer to \url{https://github.com/AMALIA-LLM/arquivo_processing} and \url{https://github.com/AMALIA-LLM/datatrove-amalia}}

\end{abstract}

\section{Introduction}
The success of modern Large Language Models (LLMs) relies heavily on the availability of massive, diverse, and high-quality text corpora. While general web scrapes like Common Crawl provide an abundant source of raw text, turning these uncurated dumps into high-quality training data requires sophisticated, multi-stage processing pipelines. For dominant, high-resource languages like English, extensive investments have yielded highly optimized, open-source recipes. However, developing comparable resources for mid-to-lower-resource languages, and specifically localized regional variants, remains a major hurdle for the natural language processing community. This challenge is particularly visible in the context of European Portuguese (PT-PT): while Portuguese is a global pluricentric language spoken globally by hundreds of millions of people, it is divided primarily into European Portuguese and Brazilian Portuguese, alongside distinct African and Asian varieties. However, when it comes to data representativeness, it is overwhelmingly dominated by the Brazilian Portuguese (PT-BR) variety. 

Standard Web-scraping and language identification filters often fail to adequately distinguish between PT-PT and PT-BR variants, and aggressively discard valid PT-PT content due to rigid heuristic boundaries.
In this work, we present a highly efficient, end-to-end data processing pipeline built on top of the open-source Datatrove~\cite{penedo2024datatrove} library to build a high-quality, production-ready PT-PT Web corpus. Our pipeline processed several web collections from Arquivo.pt, the Portuguese Web Archive,\footnote{\url{https://arquivo.pt/collections/}} representing approximately 411 terabytes (TB) of raw WARC (Web ARChive) data spanning from 1997 to 2024.

\begin{figure*}[t]
    \centering
    \includegraphics[
        width=\linewidth,
        trim={0cm 0cm 0cm 0cm},
        clip
    ]{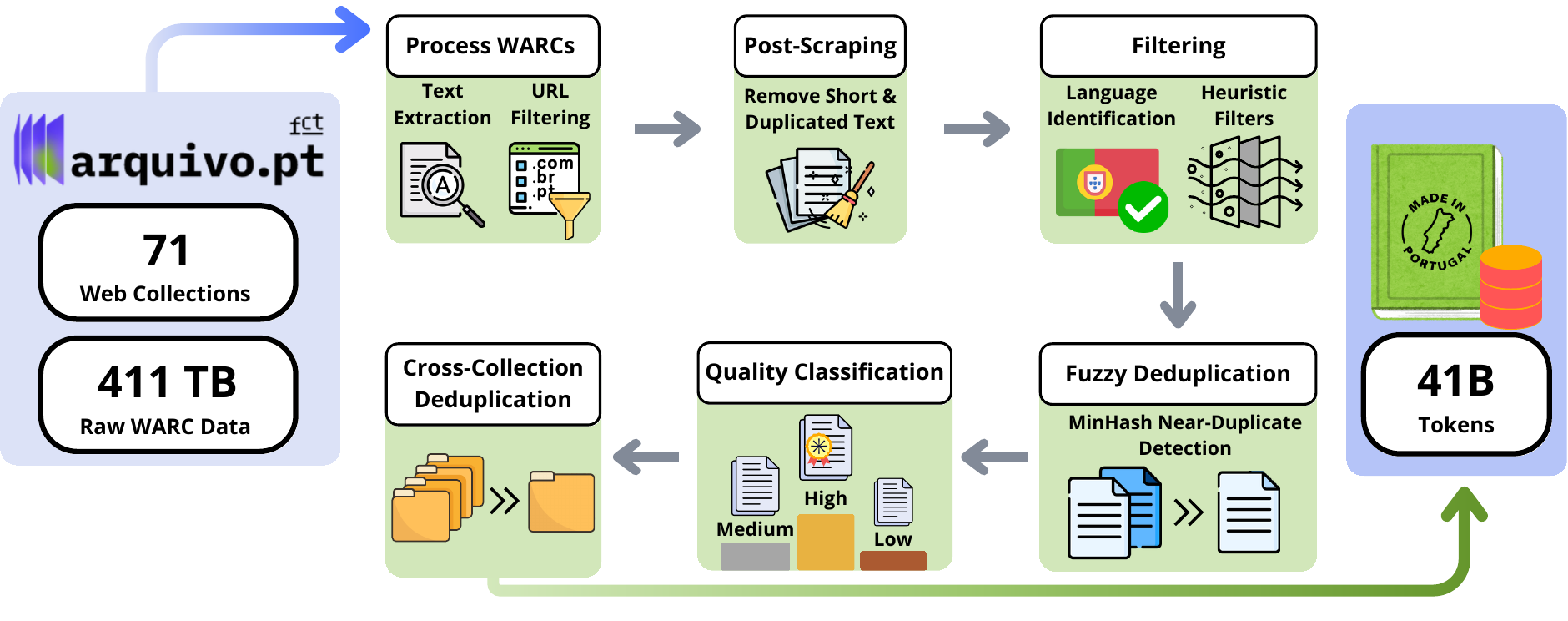}
    \caption{Overview of the developed data processing pipeline. Starting from 71 collections totaling 411 TB of raw WARC data, the pipeline applies six sequential curation stages, yielding a final corpus of 41 billion tokens of high quality European Portuguese text.}
    \label{fig:pipeline-overview}
\end{figure*}

In our pipeline, depicted in Figure~\ref{fig:pipeline-overview}, we design and implement a custom post-scraping mechanism that cleans and structures raw scraped text before passing it to downstream heuristic quality filters. This component removes short, fragmented lines and local duplicate boilerplate text within documents.
Our development experiments demonstrate that post-scraping substantially improves the effective quality of Web documents, rescuing valuable, linguistically rich data that standard filters would normally reject. This intervention alone yields a 19.04\% increase in the final number of retained high-quality documents without compromising the rigorous standards required for LLM training. The resulting pipeline combines geographical domain filtering, neural language identification, heuristic text filters adapted from FineWeb2~\cite{penedo2025fineweb2pipelinescale}, custom weighted MinHash fuzzy deduplication, neural quality classification via EuroFilter~\cite{eurollm9btech}, and cross-collection deduplication. The result is a robust, clean, and representative PT-PT corpus specifically optimized to power the mid-training and pre-training phases of next-generation language models.

\section{Related Work}
\paragraph{Web Data Processing Pipeline.}
The curation of large-scale Web corpora for LLM training has become a well-established research area, with pipelines typically built around CommonCrawl,\footnote{\url{https://commoncrawl.org}} combining filtering and deduplication steps~\cite{penedo2023refinedwebdatasetfalconllm}.
FineWeb~\cite{penedo2024finewebdatasetsdecantingweb} represents one of the major contributions to the community, releasing a 18.5-trillion-token dataset alongside Datatrove~\cite{penedo2024datatrove}, a modular data processing library, while also applying and adapting quality and repetition filters from MassiveText~\cite{rae2022scalinglanguagemodelsmethods} and C4~\cite{JMLR:v21:20-074}. 

Deduplication is equally important, as training on deduplicated data consistently improves model quality and reduces memorization~\cite{lee-etal-2022-deduplicating}. Deduplication can be done at the URL level, through exact or semantic matching~\cite{abbas2023semdedupdataefficientlearningwebscale}, or through MinHash-based near-duplicate detection~\cite{666900}, which has become the standard approach at scale. Some approaches apply heuristics to lower-quality documents only to avoid discarding high-quality content~\cite{su2025nemotroncctransformingcommoncrawl}, while others complementary apply model-based filters~\cite{henriksson-etal-2025-finerweb}.

\paragraph{Multilingual Web Corpora.}
Several efforts have been made to build high-quality multilingual Web datasets~\cite{xue-etal-2021-mt5, abadji-etal-2022-towards, NEURIPS2022_ce9e92e3,nguyen-etal-2024-culturax, ali-etal-2025-judging}, with recent work showing that language-specific pipelines are crucial to guarantee the best quality~\cite{datologyai2026uberwebinsightsmultilingualcuration,burns-etal-2026-aleph,ICLR2025_cde43c5d, penedo2025fineweb2pipelinescale}. Despite these advancements, multilingual datasets often fail to represent each language thoroughly, blurring regional variants and language-specific characteristics, a limitation that motivates language- and variant-specific corpus efforts.

\begin{figure*}[t]
    \centering
    \includegraphics[
        width=\linewidth,
        trim={0.2cm 0.2cm 0.2cm 0.2cm},
        clip
    ]{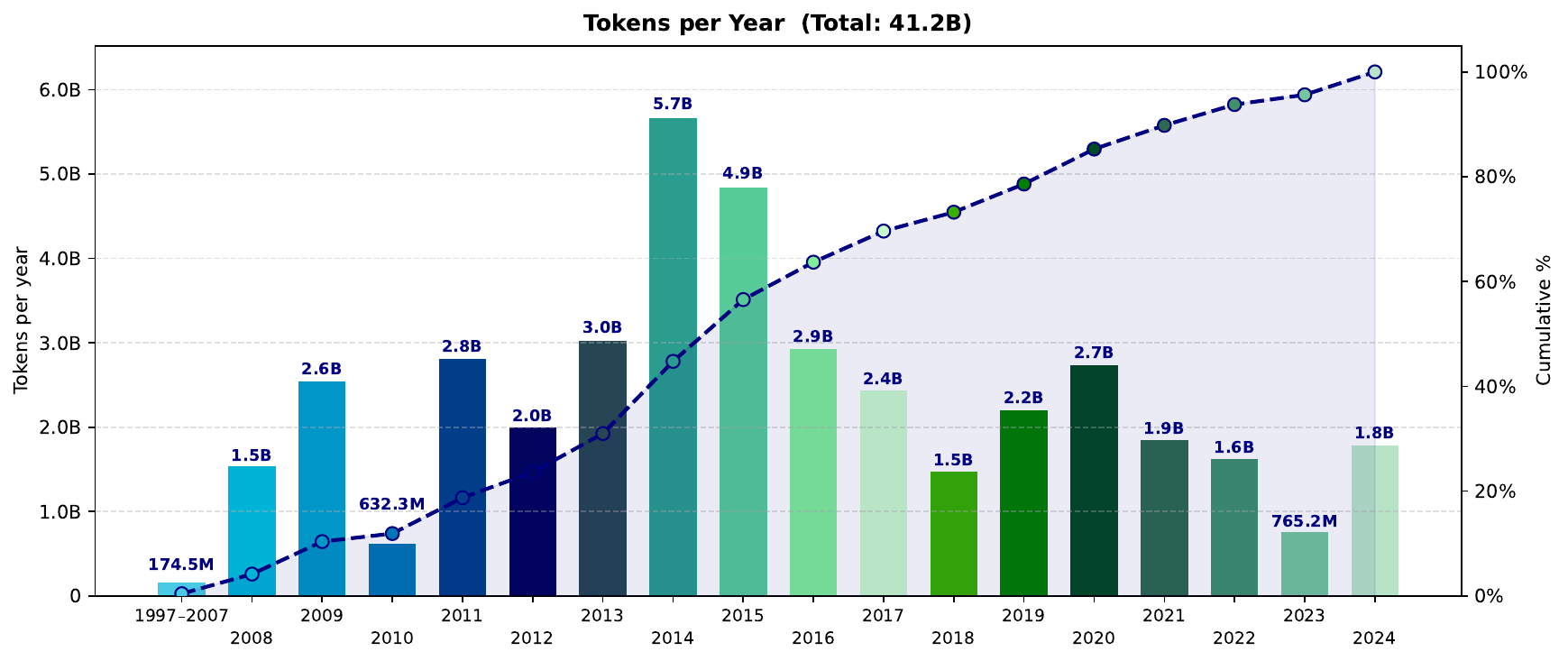}
    \caption{Token distribution across crawl years. The corpus spans 1997–2024, with peak coverage in 2014–2015. The dashed line shows the cumulative proportion of total tokens.}
    \label{fig:year_donut}
\end{figure*}

\paragraph{European Portuguese Language Resources}

While most multilingual corpora include Portuguese data, they treat it as a single variant despite significant lexical, grammatical, and orthographic differences between European and Brazilian Portuguese~\cite{preda-etal-2024-across}, with the latter being systematically over-represented due to its much larger Web presence~\cite{almeida2025buildinghighqualitydatasetsportuguese}. GlórIA~\cite{gloria_llm} introduced a European Portuguese decoder-based LLM pre-trained on a large PT-PT corpus. This corpus includes an Arquivo.pt subset derived from approximately 7~TB of raw WARC data, filtered using basic heuristic cleaning and exact-duplicate removal, resulting in 1.5 million documents (0.8 billion tokens) of `.pt' news websites only. More recently, AMALIA~\cite{simplicio-etal-2026-amalia}, the first fully open LLM for European Portuguese, was trained on a preliminary version of the corpus introduced in this work, which was made available to the AMALIA team prior to publication.

Recent benchmark developments~\cite{tavares-etal-2026-pheb, ferreira-etal-2026-p3b3, vieira-etal-2026-alba, baucells-etal-2025-iberobench} further reflect the growing PT-PT research ecosystem. Despite this progress, no large-scale, openly documented pipeline exists for processing Web archive data targeting European Portuguese specifically, a gap that this work addresses.

\begin{table*}[h]
\centering
\adjustbox{max width=\linewidth}{%
\begin{tabular}{lllrrl} \toprule
\textbf{Name} & \textbf{Type}  & \textbf{Update Rate} & \textbf{\#Tokens} & \textbf{\#URL} & \textbf{Description}                                \\ \midrule
AWP           & General Web    & Quarterly            & 36\,827M    &   20\,667k  & Complete crawls of the Portuguese Web            \\
MAWP          & Media          & Monthly              & 85M         &    84k & Monthly crawls of selected websites  \\
FAWP          & Media          & Daily                & 3\,702M     &    3\,438k  & Websites which publish new daily content        \\
EAWP          & Special Crawls & Varied               & 43M        &     31k & Selected pages about a given theme or event         \\
RAQ           & High-Quality   & Varied               & 175M       &     140k & Selected websites carefully curated    \\
Other         & Donations      & Varied               & 340M        &    199k & Collections archived/donated by users or institutions    \\ \bottomrule
\end{tabular}}
\caption{Summary of Arquivo.pt collection types processed in this work. AWP dominates the corpus by volume (36.8B tokens, 89\% of the total), while smaller curated collections such as RAQ and EAWP contribute higher-quality but sparser content.}
\label{tab:collections_summary}
\end{table*}

\section{Corpus Creation Methodology}
\subsection{Data Collection}
\label{sec:data-collection}

Our corpus is composed of publicly available data from Arquivo.pt, the Portuguese Web Archive, which has been continuously crawling and preserving Portuguese Web content since 1996, while respecting the Robots Exclusion Protocol.\footnote{\url{https://sobre.arquivo.pt/en/help/crawling-and-archiving-web-content/\#qe-faq-2407}}
We deployed a dedicated server within Arquivo.pt's infrastructure, enabling direct transfer and local processing of raw WARC archives, achieving a roughly 20$\times$ reduction in processing time relative to querying their public API,\footnote{\url{https://arquivo.pt/api}} due to bandwidth limitations and throttling.

Combining the data collected across the phases of this work, we processed 71 collections representing roughly 411 TB of raw WARC archives available in Arquivo.pt's database. These collections span a wide variety of content types, from news to blogs and the general Web, and were crawled between 1997 and 2024, ensuring that we respect an embargo of one year,\footnotemark as seen in Figure~\ref{fig:year_donut} (collection selection details are in Appendix~\ref{app:col-selection}).

\footnotetext{Arquivo.pt follows a one year embargo policy, to avoid drawing traffic away from original content publishers. }

The collections we processed were part of three main categories: frequently crawled content (FAWP), general Web content (AWP), and other smaller collections with high-quality content.  Depending on the type of collection, the crawling frequency ranges from daily to quarterly, with some collections dedicated to special events and/or selected websites. Among the available collections, we have chosen a total of 71 collections, according to their time span, spacing in time, and crawling frequency, towards minimizing duplicates while still guaranteeing content comprehensiveness over the full period. Table~\ref{tab:collections_summary} summarizes the proportions and content of each collection type processed.

\subsection{Data Processing}
\label{subsub:data_processing}

We process the raw Web archive data from Arquivo.pt, using the WARC archives collected, into a high-quality European Portuguese corpus through a multi-stage pipeline built on top of Datatrove, an open-source library for large-scale text processing. Our pipeline takes inspiration from FineWeb2~\cite{penedo2025fineweb2pipelinescale}, adopting its threshold configuration for Portuguese,\footnote{\url{https://github.com/huggingface/fineweb-2/blob/main/configs/por_Latn.yml}} but extending its approach with custom processing components and implementation improvements.  Our overall pipeline consists of six stages that each WARC collection is submitted to: (1) WARC Processing, (2) Text Post-Scraping, (3) Data Filtering, (4) MinHash Deduplication, (5) Quality Classification, and (6) Cross-Collection Deduplication. An overview is shown in Figure~\ref{fig:pipeline-overview}.

\subsubsection{WARC Processing}
\label{subsub:warc_processing}

The first stage processes the raw WARC archives, using Datatrove's built-in \textit{WarcReader}, which are submitted to two URL-level filters before any text extraction. First, we remove every document with URLs belonging to `.br' domains or containing similar variants (such as `/pt-br'), in order to reduce the presence of Brazilian Portuguese text and focus on European Portuguese. Second, we remove URLs matching a blacklist of known NSFW domains and banned keywords. We deliberately do not restrict to `.pt' domains alone, as this would discard a large proportion of valid European Portuguese content hosted under other top level domains.

We proceed with the text extraction from the HTML content in the WARC files using a custom Trafilatura-based~\cite{barbaresi-2021-trafilatura} block that extends Datatrove's built-in implementation to additionally preserve the page title as metadata alongside the extracted text. We find that this extra metadata provides richer document context for downstream use without decreasing extraction efficiency.

\subsubsection{Text Post-Scraping}
\label{subsubsec:text_postscraping}

One of the main contributions of this work is a custom \textit{post-scraping} block, which improves the effective quality of the documents. This improvement prevents some documents from being removed by the downstream heuristic filters, without worsening the final quality of the dataset. This represents an increase of 19.04\% in the final number of documents obtained from our test collection (more details in Section~\ref{sec:post-scraping_experiments}). We remove short lines, i.e., with fewer than 10 alphabetic characters, and duplicate lines within a document, retaining only a single instance of each repeated line.

\subsubsection{Filtering}
\label{subsubsec:filtering}

The extracted and processed text is then submitted to quality filters to discard low-quality documents. First, each document is passed through a \textit{Language Filter} using the GlotLID model~\cite{Kargaran_2023}. Documents where the estimated probability of European Portuguese falls below a threshold of 0.799 (consistent with the FineWeb2 configuration for Portuguese) are discarded.

For the heuristic quality filters, we start with the \textit{Gopher Repetition Filter}~\cite{rae2022scalinglanguagemodelsmethods} to remove documents with excessive repetition, removing documents with a ratio of duplicated n-grams following two different criteria: for short n-grams (2–4), we apply a top n-gram criterion that flags documents where the most frequent repeated n-gram accounts for a disproportionate share of characters; for longer n-grams (5–10), we apply a duplicate n-gram criterion over all repeated n-gram occurrences. Next, we apply the \textit{FineWeb Quality Filter}, removing documents with excessive lines ending without appropriate punctuation and with an excessive ratio of newlines to words. Finally, the \textit{Gopher Quality Filter}~\cite{rae2022scalinglanguagemodelsmethods} is applied, where documents are removed if they fall outside acceptable word count bounds, have an average word length outside the defined range, exceed limits on the proportion of hash symbols, ellipses, bullet-prefixed lines, or ellipsis-terminated lines, fall below a minimum alphabetic character ratio, or contain fewer than two Portuguese stop words from a fixed list. An example of a sample removed by each heuristic filter is shown in Appendix~\ref{app:removed-samples}. 
Following the heuristic filters, we apply a formatting step to correct possible character encoding issues, redact personal data (public IP addresses, email addresses, and phone numbers), and remove residual HTML table artifacts introduced by the scraper.

\subsubsection{Deduplication and Rehydration}
\label{sec:deduplication}

For each collection, we independently perform fuzzy deduplication using MinHash~\cite{666900}, implemented via Datatrove's four-stage deduplication pipeline: (1) computing a MinHash signature for each document, (2) identifying pairs of duplicate signatures and outputting them as buckets, (3) creating clusters of duplicates from the buckets, and (4) filtering the original documents by retaining only one representative document per cluster.

After deduplication, a \textit{Rehydrater} block assigns an upsampling weight to each retained document based on its cluster size. Documents from larger clusters receive proportionally higher weights, reflecting the implicit density of the content they represent, whereas documents from very large clusters (indicative of spam or boilerplate) have their weights capped. These weights are stored in the document metadata to support optional data rehydration during training. Besides these weights, our custom Datatrove implementation also lets users keep all duplicate documents and only annotate their clusters' ids and sizes, to allow for more cluster filtering techniques besides random sampling, as Datatrove defaults to.

\subsubsection{Quality Classification}

Following per-collection deduplication, all collections are jointly processed by the EuroFilter quality classification model~\cite{eurollm9btech}. With the scores given by this classifier, we discarded the low-quality entries and proceeded to the following pipeline stage only with the remaining ones. Unlike the other stages, this step is run outside of Datatrove.

\subsubsection{Cross-Collection Deduplication}

The per-collection deduplication in Section~\ref{sec:deduplication} does not remove duplicates that appear across collections. Since different Arquivo.pt collections often crawl the same Web domains at different points in time, they may overlap substantially in content. Thus, a second fuzzy deduplication pass, using an identical procedure, is performed across all collections jointly, applied to the output of the quality classification stage. This step yields the final corpus used for language model training.

\subsection{Hardware and Processing Configuration}

As mentioned in Section~\ref{sec:data-collection}, we deployed a dedicated server in Arquivo.pt's infrastructure, achieving a 20$\times$ reduction in total processing time compared to querying their public API. This dedicated server has two AMD CPUs with 32 cores each, one JBOD storage array with 1.2 PB, and a 20 Gbps network connection. WARC processing was performed on this dedicated server and took three weeks, while filtering and deduplication were performed on the MareNostrum 5 supercomputer hosted at the Barcelona Supercomputing Center, using seven compute nodes, and together took one and a half weeks to complete. All three stages used Datatrove, submitting jobs as SLURM job arrays with its predefined parameters: WARC processing ran with 10,000 tasks across 32 workers, using 2 CPUs per task and 4GB of memory per CPU; data filtering ran with 280 tasks across 128 workers, using 6 CPUs per task and 2GB of memory per CPU; deduplication ran with 100 tasks across 100 workers, using 8 CPUs per task and 2GB of memory per CPU.

\section{Data Curation Results}

We validate the design decisions in our pipeline through experiments on a single collection (FAWP21), composed primarily of Portuguese news media, frequently updated websites, and government and public body pages, crawled between April and June 2015.

\subsection{Post-Scraping Steps}
\label{sec:post-scraping_experiments}

\begin{figure}
    \centering
    \includegraphics[
        width=\linewidth,
        trim={0.2cm 0.2cm 0.2cm 0.2cm},
        clip
    ]{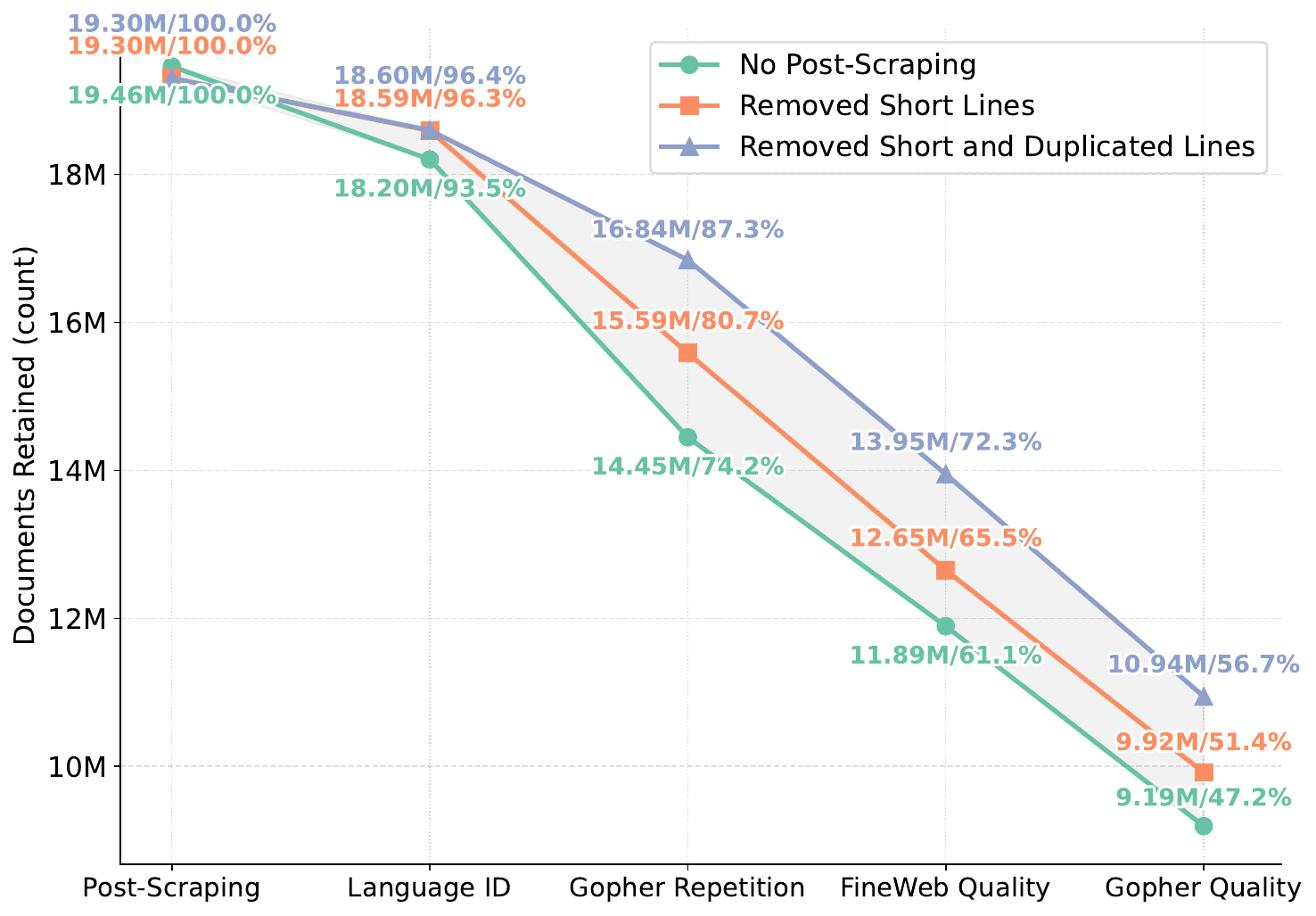}
    \caption{Comparison of document retention through the filtering stages between not including any of the post-scraping operations, including only the short line removal operation, and including the short line and duplicated lines removal operations.}
    \label{fig:post_scraping_ablation}
\end{figure}

We remove lines with fewer than 10 characters and duplicate lines within each document before heuristic filtering. To understand the contribution of each added step, we compare three configurations: no post-scraping (baseline), short line removal only, and both steps combined.

Performing both steps consistently increases the number of documents remaining after the full filtering pipeline. Starting from a baseline of 9.19 million retained documents, adding short line removal yields 9.92 million documents (+7.94\%), while further adding duplicate line removal yields 10.94 million documents (+19.04\% over baseline).
This yield improvement does not come at the expense of quality: post-scraping improves the quality of documents which would otherwise be rejected. Every document, regardless of post-scraping, still passes through the same downstream filtering pipeline. Each step rescues more content which would have been discarded by the heuristic filters in its raw scraped form, as shown in Figure~\ref{fig:post_scraping_ablation}. In our test collection, only 47.2\% of documents survive to the final Gopher Quality stage without post-scraping, compared to 56.7\% with both steps applied. Examples of rescued documents before and after being subject to post-scraping can be seen in Appendix~\ref{app:rescued_samples}.
We also experimented removing residual HTML tags from the extracted text, but found that many were part of legitimate technical content such as HTML tutorials, where their presence is meaningful. Consequently, we utilize the text scraper's native tag cleanup feature

\subsection{Model-Based Quality Pre-Filtering}

Motivated by Nemotron-CC~\cite{su2025nemotroncctransformingcommoncrawl}, we evaluated whether applying quality classification before heuristic filtering, exempting high-quality documents, could improve our pipeline. We thus applied the EuroFilter quality classifier~\cite{eurollm9btech} to our test collection prior to filtering.
The results showed negligible benefit, as this pipeline retained 0.4\% more documents than heuristic filtering alone. 
Given that this step significantly increases the processing time for a single collection, the marginal yield improvement does not justify the added computational cost. As a result we used the simpler heuristic-only filtering approach.

\begin{table}[t]
\centering
\begin{tabular}{lrr}
\toprule
\textbf{Stage} & \textbf{Documents}\\
\midrule
Raw WARC data (\S\ref{sec:data-collection})         & 5\,392\,417\,365 \\
After URL filtering (\S\ref{subsub:warc_processing})   & 3\,440\,781\,193 \\
After text extraction (\S\ref{subsub:warc_processing}) & 1\,653\,754\,362 \\
After heuristics (\S\ref{subsubsec:filtering})      & 731\,048\,214 \\
After deduplication (\S\ref{sec:deduplication})   & 147\,699\,937 \\
\bottomrule
\end{tabular}
\caption{Document counts at each stage of the processing pipeline across all 71 Arquivo.pt collections before cross-collection deduplication and quality classification}
\label{tab:pipeline_reduction}
\end{table}

\subsection{Final Corpus Statistics}
\label{sec:corpus-results}

This pipeline resulted in a high-quality European Portuguese Web corpus, enabling pre-training or mid-training of PT-PT-focused language models. Table~\ref{tab:pipeline_reduction} summarizes the filtering efficiency at each stage of the processing pipeline over all 71 collections. Starting from approximately 5.4 billion raw entries in the WARC archives, URL filtering reduces this to 3.4 billion entries by removing Brazilian Portuguese and NSFW domains. Text extraction further reduces the count to 1.7 billion entries, after which heuristic filtering removes the majority of low-quality documents, leaving 731 million entries. Single-collection MinHash deduplication then produces a dramatic reduction to 148 million entries, an \textasciitilde{}80\% reduction from the previous step and \textasciitilde{}97\% from the raw input, reflecting the high degree of content repetition inherent in a Web archive crawled over multiple years, particularly in news-heavy collections where pages are crawled daily.

After discarding low-quality content through quality classification, the remaining documents are subject to a cross-collection deduplication stage. This results in 24.6M documents and 41B tokens (calculated using the GPT-2 tokenizer~\cite{radford2019language}, as are all token counts in this section). These processing steps reduce the corpus token count from 176B tokens (after per-collection deduplication) to 41B tokens, a 76.6\% reduction, reflecting substantial content overlap across collections and highlighting the importance of this processing step. When comparing with the Arquivo.pt data included in GlórIA's training corpus, this represents a 16.4$\times$ increase in document count and a 51.3$\times$ increase in token count, large enough for Arquivo.pt alone to serve as a primary pretraining source rather than a minor supplementary one, as it was in GlórIA's corpus, as summarized in Table~\ref{tab:gloria-ours-comparison}.

\begin{figure*}[t]
    \centering
    \includegraphics[
        width=0.9\linewidth,
        clip
    ]{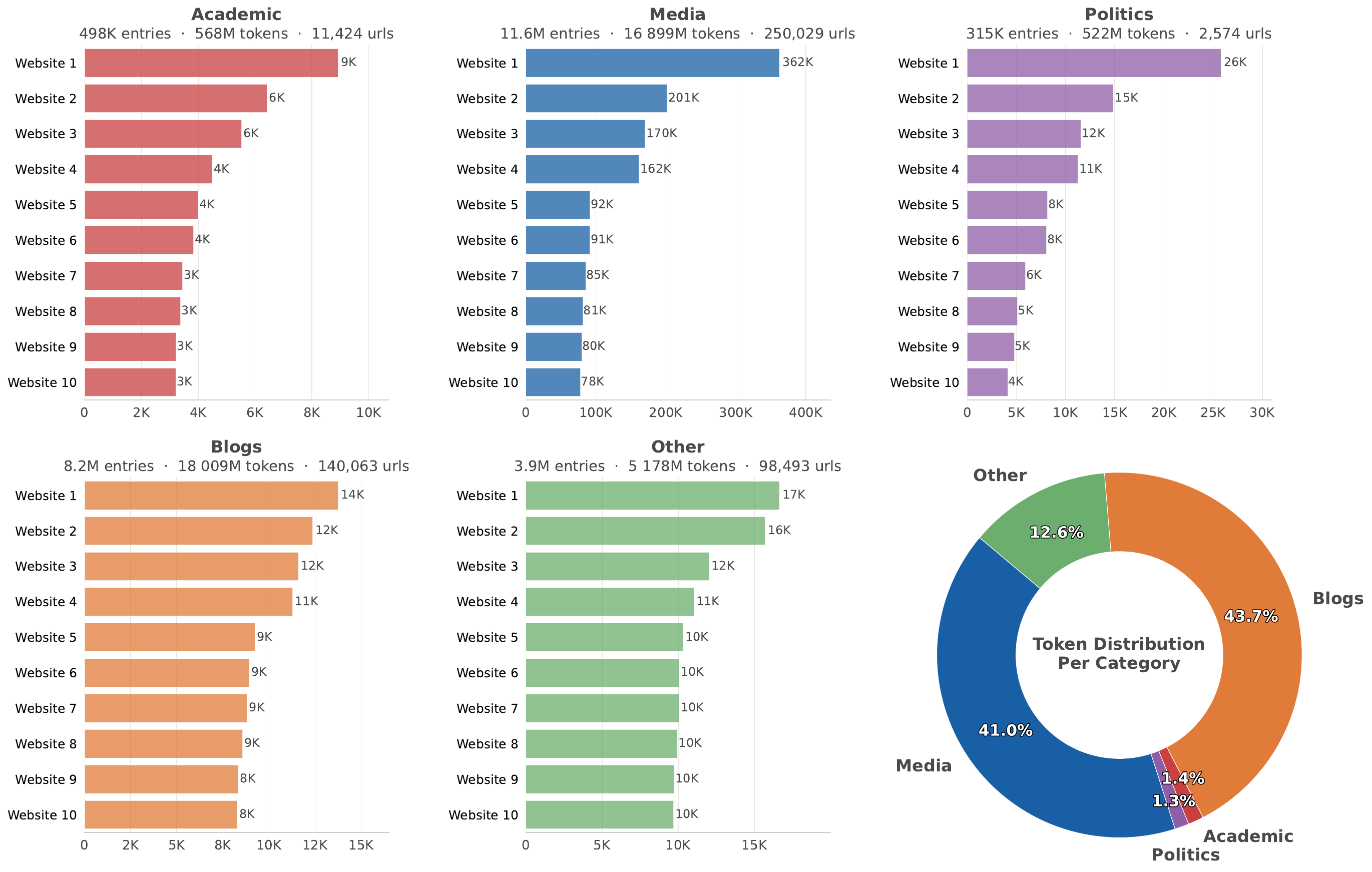}
    \caption{Top 10 domains by entry count across content categories, with overall token distribution by category (bottom right). The corpus is dominated by media and blogs (41\% and 43.7\% of tokens, respectively), reflecting the strong presence of professional Portuguese-language publishing in Arquivo.pt.}
    \label{fig:entries-per-category}
\end{figure*}

\begin{table}[t]
\centering
\begin{tabular}{lcc} 
\toprule
            & \textbf{GlórIA} & \textbf{Ours}  \\ 
\midrule
\#Documents & 1.5M            & 24.6M         \\
\#Tokens    & 0.8B            & 41B           \\
Scope       & News only       & All domains    \\
\bottomrule
\end{tabular}
\caption{Comparison between the Arquivo.pt data used in GlórIA vs our approach.}
\label{tab:gloria-ours-comparison}
\end{table}

\begin{figure}[t]
    \centering
    \includegraphics[
        width=\linewidth,
        trim={0.2cm 0cm 0.2cm 0cm},
        clip
    ]{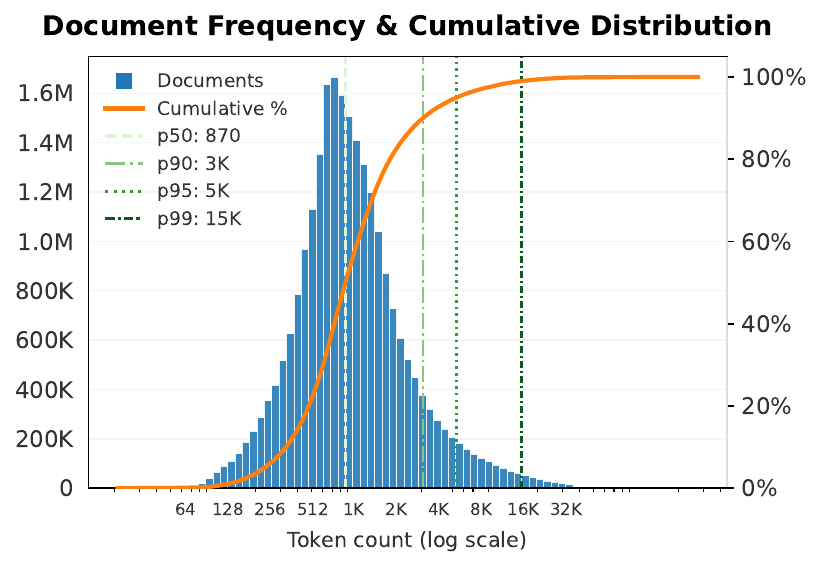}
    \caption{Document token-length distribution of the processed Arquivo.pt collections.}
    \label{fig:token-length}
\end{figure}

The final corpus covers a wide variety of content areas, with news and blogs dominating; Figure~\ref{fig:entries-per-category} shows the 10 most common domains per category and the category distribution of the corpus. Figure~\ref{fig:token-length} shows the cumulative token length distribution of the final corpus, with over 90\% of documents containing fewer than 4k tokens.

\section{Experiments and Results}
\label{sec:corpus-results}

\subsection{Filter Stage Quality Ablations}

\begin{table*}[ht]
\centering
\scriptsize
\setlength{\tabcolsep}{3pt}

\resizebox{\textwidth}{!}{%
\begin{tabular}{l cccccc cccccc cccccc}
\toprule
 & \multicolumn{6}{c}{Consecutive prefix words} & \multicolumn{6}{c}{BLEU-3} & \multicolumn{6}{c}{BLEU-4} \\
\cmidrule(lr){2-7} \cmidrule(lr){8-13} \cmidrule(lr){14-19}
Model & mean & p75 & p90 & p95 & p99 & max & mean & p75 & p90 & p95 & p99 & max & mean & p75 & p90 & p95 & p99 & max \\
\midrule
AMALIA-9B & 0.696 & 1.0 & 2.0 & 3.0 & 6.0 & 26.0 & 0.051 & 0.068 & 0.104 & 0.138 & 0.219 & 0.391 & 0.031 & 0.038 & 0.067 & 0.098 & 0.169 & 0.326 \\
EuroLLM-9B & 0.704 & 1.0 & 2.0 & 3.0 & 7.0 & 26.0 & 0.048 & 0.062 & 0.098 & 0.134 & 0.203 & 0.474 & 0.028 & 0.034 & 0.062 & 0.092 & 0.15 & 0.462 \\
Llama-3.1-8B & 0.622 & 1.0 & 2.0 & 3.0 & 5.0 & 18.0 & 0.045 & 0.058 & 0.094 & 0.124 & 0.197 & 0.256 & 0.026 & 0.031 & 0.056 & 0.082 & 0.146 & 0.213 \\
OLMo-2-7B & 0.449 & 1.0 & 1.0 & 2.0 & 4.0 & 18.0 & 0.036 & 0.046 & 0.072 & 0.096 & 0.166 & 0.307 & 0.021 & 0.024 & 0.042 & 0.062 & 0.13 & 0.276 \\
\bottomrule
\end{tabular}
}

\caption{Memorization-eval statistics per model: means, percentiles and maximum.}
\label{tab:memorization}
\end{table*}

To empirically validate the effect of each filtering stage on document quality, we analyze the next-token prediction behavior of a set of language models on documents removed at each stage of the heuristic filtering pipeline. For each filter, we sample 10,000 documents from the removed set, selecting the worst-case examples according to the filter metric that caused their removal (e.g., the documents with the highest duplicate n-gram fraction for Gopher Repetition), while preserving the original proportion of filter reasons.  We additionally sample 10,000 documents uniformly at random from the pipeline output after filtering, which serves as our quality baseline. These sets are deduplicated before sampling to ensure diversity. For each document, we compute bits-per-byte (bpb) and top-1 next-token prediction accuracy using four base language models of comparable size: AMALIA 9B~\cite{simplicio-etal-2026-amalia}, a PT-PT-focused model mid-trained in our collection; EuroLLM 9B~\cite{eurollm9btech}, the base model on top of which AMALIA was trained; Llama 3.1 8B~\cite{grattafiori2024llama}, an established multilingual model; and OLMo 2 7B~\cite{OLMo20242O2}, an English-only model included as a control.

\begin{figure}
    \centering
    \includegraphics[
        width=\linewidth,
        trim={0.2cm 0.5cm 0.2cm 0.2cm},
        clip
    ]{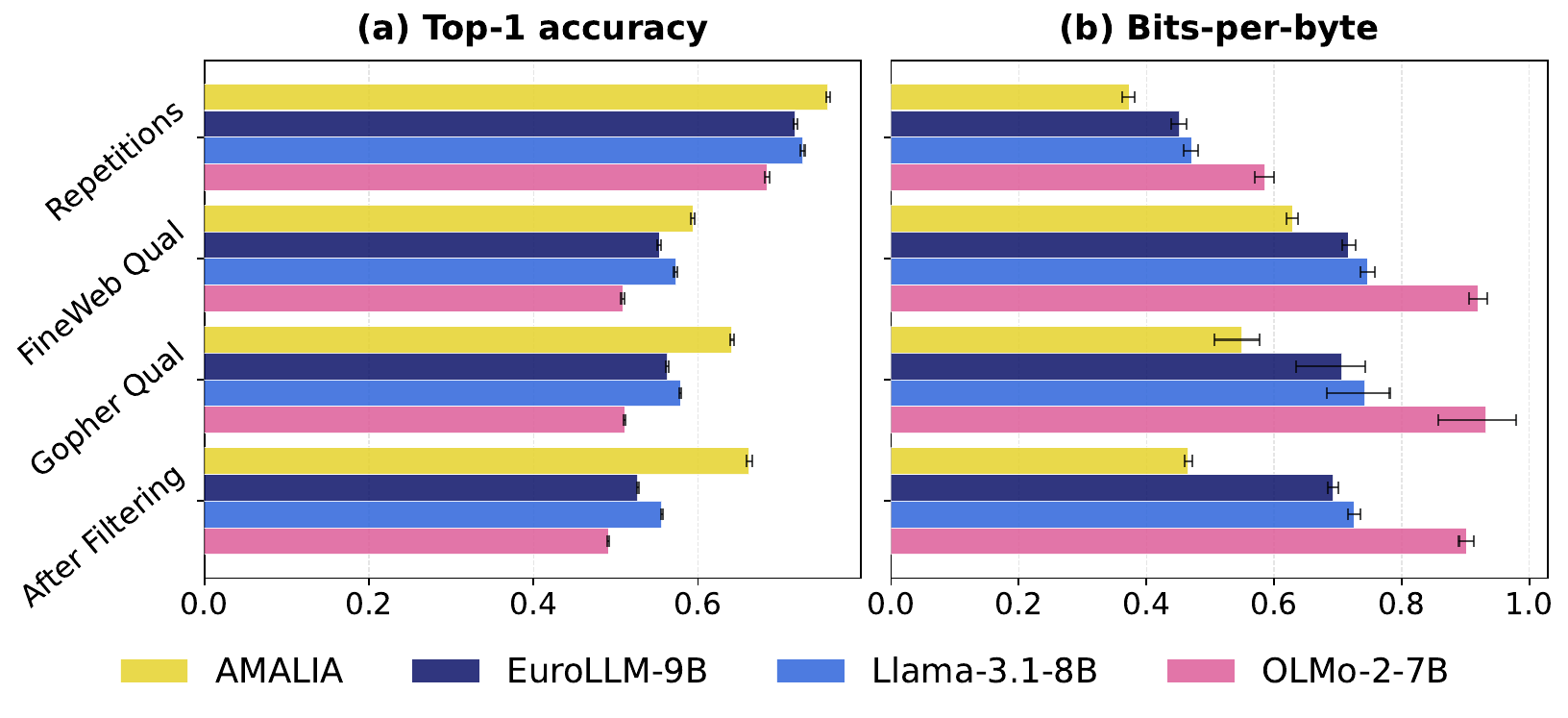}
    \caption{Model comparison on (a) top-1 next-token prediction accuracy and (b) bits-per-byte (lower=better) on the removed samples by each filter stage and the final output.}
    \label{fig:ntp-evolution}
\end{figure}

The results, shown in Figure~\ref{fig:ntp-evolution}, reveal a consistent pattern across all four models. Documents removed by the FineWeb Quality and Gopher Quality filters are systematically harder to predict than the final output, with higher bits-per-byte across all models, confirming that these filters remove genuinely low-quality text that deviates from the natural language patterns learned during pre-training. However, the documents removed at the Gopher Repetition stage have the lowest bits-per-byte and highest top-1 accuracy. This is expected, since repetitive documents are inherently predictable and a language model that has learned any recurring pattern will assign high probability to its continuations, scoring better than the other stages’ samples.

AMALIA shows markedly lower bits-per-byte and higher top-1 accuracy on the output set, reflecting the effect of its mid-training on Arquivo.pt data from this exact collection. However, EuroLLM's bits-per-byte on the output is comparable to or lower than its performance on the Gopher Quality removed set, suggesting that EuroLLM's general multilingual training does not specifically favour high-quality European Portuguese web text over low-quality text from the same domain. 

OLMo-2, an English-only model, shows the highest bits-per-byte and lowest top-1 accuracy accross all stages, reflecting its lack of exposure to Portuguese during pre-training. Despite this, it mirrors the ordering observed in the multilingual models. This suggests that these patterns reflect intrinsic properties of the text rather than any model's familiarity with Portuguese.

\subsubsection{Quality Thresholds Comparison}

\begin{figure}
    \centering
    \includegraphics[
        width=\linewidth,
        trim={0.2cm 0.2cm 0.2cm 0.2cm},
        clip
    ]{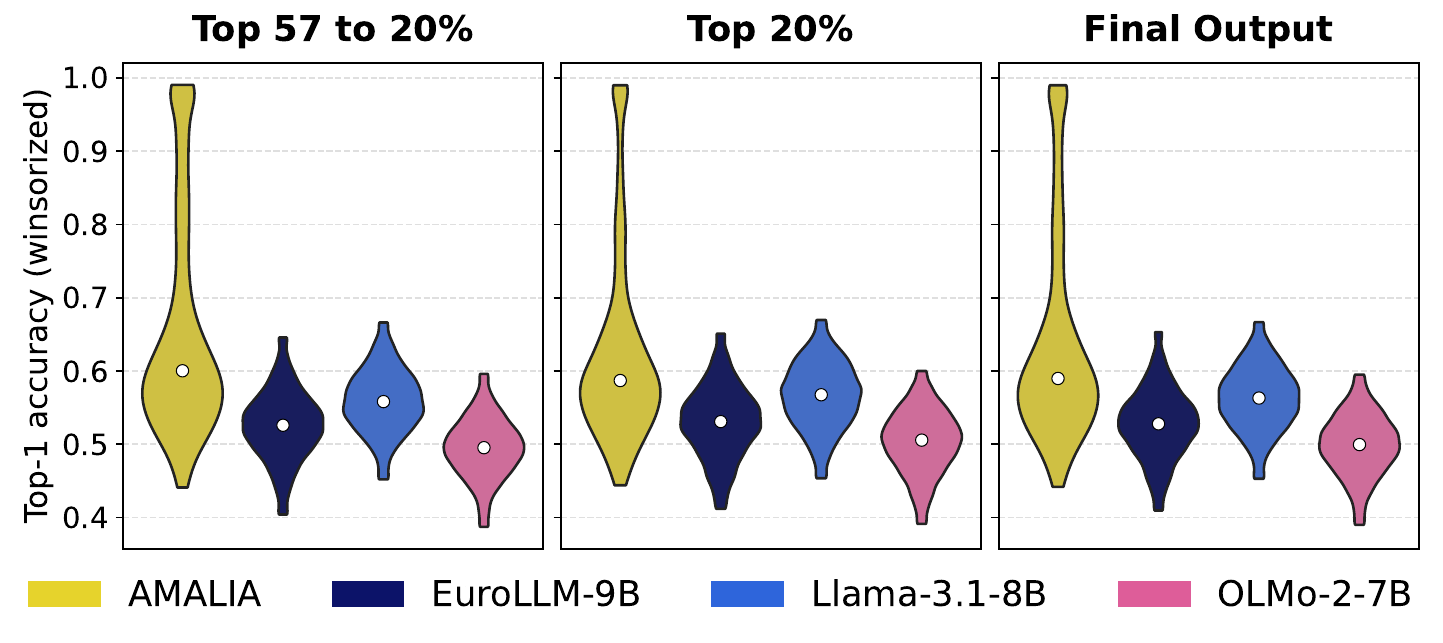}
    \caption{Quality thresholds comparison in the next-token prediction task.}
    \label{fig:quality-splits-comparison}
\end{figure}

Applying our quality filter followed by cross-collection deduplication on the medium and high-quality documents, leaves us with 57\% of the original documents. To assess the impact of our quality threshold, we experimented running our next-token prediction evaluation at different splits of our final corpus: top 57 to 20\%, which we'll call medium-quality; top 20\%, which we'll call high-quality; and the final output, consisting of both medium- and high-quality splits. Figure~\ref{fig:quality-splits-comparison} shows that AMALIA achieves slightly higher top-1 accuracy on medium-quality documents than on high-quality ones, with averages of 0.68 and 0.63, respectively, likely reflecting the larger proportion of medium-quality text in its training data. This suggests that increasing the proportion of high-quality documents during training, achievable through the upsampling weights assigned by the \textit{Rehydrater} block, could improve the quality of the final corpus, and, consequently, the trained model's capabilities.

\subsection{Memorization Experiments}

\begin{figure}
    \centering
    \includegraphics[width=\linewidth]{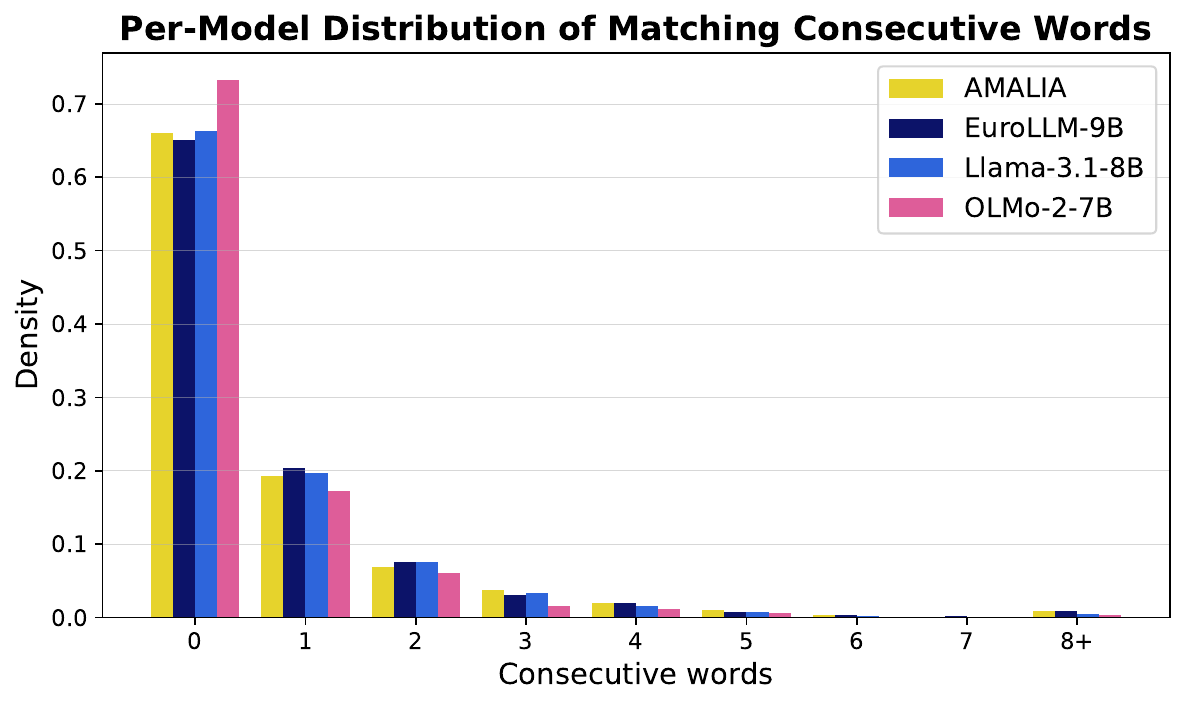}
    \caption{Per-model distribution of the count of leading consecutive words that exactly match the ground-truth continuation (n = 1,000 prompts each).}
    \label{fig:consecutive_word_hist}
\end{figure}

Language models are known to memorize and regurgitate verbatim sequences from their training data, raising privacy and copyright concerns~\cite{verbatim-memo} and inflating perplexity-based quality estimates~\cite{lee-etal-2022-deduplicating}. It is therefore crucial to understand if a model trained with this data is generalizing or memorizing the data. To evaluate this, we adopt a prefix-continuation protocol: 1) we sample documents from the corpus, 2) truncate each sample to at most 4{,}096 tokens, 3) split them in half, and 4) prompt the model with the first half. We then greedy-decode a continuation of the same length as the held-out second half and measure the overlap between the generated and the true continuation. Our primary metric is \emph{consecutive words}: the length of the longest word sequence, starting from the first generated token, that the model reproduces exactly from the true continuation. This directly captures verbatim memorization by measuring how far the model can extend a document word-for-word before diverging. We complement this with \emph{BLEU-3} and \emph{BLEU-4}, which quantify n-gram precision at orders 3 and 4 between the generated and reference continuations. For each metric we report the mean, maximum, and upper percentiles across samples, summarized in Table~\ref{tab:memorization}; per-model distributions of \emph{consecutive words} are shown in Figure~\ref{fig:consecutive_word_hist}. The results indicate that memorization is minimal: over 99\% of samples yield six or fewer consecutive matching words, and BLEU-4 scores remain below 0.169, confirming that the models rarely reproduce verbatim spans from the training data. The few exceptions are still small as illustrated in Table~\ref{tab:mem-examples}. Notably, AMALIA's memorization metrics remain comparable to those of EuroLLM, the base model on top of which it was mid-trained, suggesting that the additional training on this corpus did not lead to meaningful memorization beyond what is already attributable to general language patterns and potential overlap with other pretraining sources.

\subsection{Brazilian Portuguese Contamination}

Due to the significant overlap between European and Brazilian Portuguese variants, it is difficult to filter out all PT-BR content without losing high-quality PT-PT text. Our URL-based filtering approach does not exclude generic top-level domains such as \texttt{.com} and \texttt{.net}, which may include content in either Portuguese variant. We therefore estimated PT-BR contamination in the final corpus using two independent, well-established language-variety classifiers. The \texttt{duarteocarmo/fasttext-euptvid} classifier~\cite{euptvid2026} labeled 86.4\% of the corpus as PT-PT and 13.6\% as PT-BR, while the \texttt{bastao/PeroVaz\_PT-BR\_Classifier}\footnote{\url{https://huggingface.co/bastao/PeroVaz_PT-BR_Classifier}} assigned 82.9\% to PT-PT and 17.1\% to PT-BR. Averaging the estimates from the two classifiers results in 84.65\% PT-PT and 15.35\% PT-BR, indicating that PT-PT remains the predominant variant in the final corpus and providing evidence for the effectiveness of our filtering steps.

\section{Conclusion}

This paper presents an efficient pipeline for producing and curating a PT-PT corpus at scale, in order to address the low availability of text resources of the language variant when compared to PT-BR. Using the Web content preserved by Arquivo.pt, the Portuguese Web Archive, and taking advantage of a dedicated server infrastructure, we processed 411 TB of raw WARC archives, spanning from 1997 to 2024. Our pipeline introduced a novel post-scraping block, improving the overall quality of the text, leading to an increase of 19.04\% in the document count of the final corpus. In combination with URL, language, and heuristic-based filtering, model-based quality classification, and MinHash fuzzy deduplication steps, we were able to create a \textbf{41 billion tokens of high-quality European Portuguese dataset}.


\section*{Acknowledgments}
This work was supported by the AMALIA project under Measure RE-C05-i08 of the Portuguese national Programa de Recuperação e Resiliência. We also acknowledge the support of Fundação para a Ciência e Tecnologia (FCT) and the NOVA LINCS project (UID/04516/2025). We thank Arquivo.pt for providing the archived data and collaborating on setting up a dedicated server within their infrastructure, which made this work possible. Finally, we thank the Barcelona Supercomputing Center (BSC) for providing the needed computational resources.

\clearpage
\section*{Limitations} 

The corpus is topically skewed towards news media and frequently updated websites, which dominate the Arquivo.pt collections we processed. Despite its naturally high journalistic quality and comprehensive coverage of Portugal events, the resulting corpus may still underrepresent certain content types such as academic, technical, or literary text, which are less prevalent in web archive data in general. We counteracted this skewness by including more general collections (AWP) and other smaller curated collections to improve topical diversity.

The heuristic filters applied in this pipeline operate at the document level, discarding entire documents when any quality threshold is exceeded. As a result, documents containing mostly high-quality text alongside small portions of low-quality content are removed in their entirety. This conservative approach was deliberately chosen to minimise the risk of noisy or malformed text influencing LLM training, but it inevitably discards some useful content. A sentence- or paragraph-level filtering approach could recover this content at the cost of additional complexity.

\section*{Ethical Considerations}

The development of this pipeline and the collection of Arquivo.pt data were carried out in accordance with relevant ethical and legal guidelines. Following GDPR legislation, personal data was removed from the final corpus. Only publicly available information was collected, while following all copyright and licensing restrictions. The Robots Exclusion Protocol (robots.txt) directives of each crawled domain were also respected throughout the collection process. 

\bibliography{custom}

@inproceedings{gloria_llm,
  author       = {Ricardo Lopes and
                  Jo{\~{a}}o Magalh{\~{a}}es and
                  David Semedo},
  title        = {Gl{\'{o}}rIA: {A} Generative and Open Large Language Model for
                  Portuguese},
  booktitle    = {{PROPOR} 2024, Volume 1},
  pages        = {441--453},
  publisher    = {ACL},
  year         = {2024},
  url          = {https://aclanthology.org/2024.propor-1.45},
  bibsource    = {dblp computer science bibliography, https://dblp.org}
}

@misc{penedo2025fineweb2pipelinescale,
      title={FineWeb2: One Pipeline to Scale Them All -- Adapting Pre-Training Data Processing to Every Language}, 
      author={Guilherme Penedo and Hynek Kydlíček and Vinko Sabolčec and Bettina Messmer and Negar Foroutan and Amir Hossein Kargaran and Colin Raffel and Martin Jaggi and Leandro Von Werra and Thomas Wolf},
      year={2025},
      eprint={2506.20920},
      archivePrefix={arXiv},
      primaryClass={cs.CL},
      url={https://arxiv.org/abs/2506.20920}, 
}

@misc{penedo2024datatrove,
  author = {Penedo, Guilherme and Kydlíček, Hynek and Cappelli, Alessandro and Sasko, Mario and Wolf, Thomas},
  title = {DataTrove: large scale data processing},
  year = {2024},
  publisher = {GitHub},
  journal = {GitHub repository},
  url = {https://github.com/huggingface/datatrove}
}

@inproceedings{barbaresi-2021-trafilatura,
  author       = {Adrien Barbaresi},
  editor       = {Heng Ji and
                  Jong C. Park and
                  Rui Xia},
  title        = {Trafilatura: {A} Web Scraping Library and Command-Line Tool for Text
                  Discovery and Extraction},
  booktitle    = {{ACL} 2021 - System Demonstrations},
  pages        = {122--131},
  publisher    = {ACL},
  year         = {2021},
  url          = {https://doi.org/10.18653/v1/2021.acl-demo.15},
  doi          = {10.18653/V1/2021.ACL-DEMO.15},
  bibsource    = {dblp computer science bibliography, https://dblp.org}
}

@inproceedings{Kargaran_2023,
   title={GlotLID: Language Identification for Low-Resource Languages},
   url={http://dx.doi.org/10.18653/v1/2023.findings-emnlp.410},
   DOI={10.18653/v1/2023.findings-emnlp.410},
   booktitle={Findings of the Association for Computational Linguistics: EMNLP 2023},
   publisher={Association for Computational Linguistics},
   author={Kargaran, Amir and Imani, Ayyoob and Yvon, François and Schuetze, Hinrich},
   year={2023},
   pages={6155–6218} 
}

@misc{rae2022scalinglanguagemodelsmethods,
      title={Scaling Language Models: Methods, Analysis \& Insights from Training Gopher}, 
      author={Jack W. Rae and Sebastian Borgeaud and Trevor Cai and Katie Millican and Jordan Hoffmann and Francis Song and John Aslanides and Sarah Henderson and Roman Ring and Susannah Young and Eliza Rutherford and Tom Hennigan and Jacob Menick and Albin Cassirer and Richard Powell and George van den Driessche and Lisa Anne Hendricks and Maribeth Rauh and Po-Sen Huang and Amelia Glaese and Johannes Welbl and Sumanth Dathathri and Saffron Huang and Jonathan Uesato and John Mellor and Irina Higgins and Antonia Creswell and Nat McAleese and Amy Wu and Erich Elsen and Siddhant Jayakumar and Elena Buchatskaya and David Budden and Esme Sutherland and Karen Simonyan and Michela Paganini and Laurent Sifre and Lena Martens and Xiang Lorraine Li and Adhiguna Kuncoro and Aida Nematzadeh and Elena Gribovskaya and Domenic Donato and Angeliki Lazaridou and Arthur Mensch and Jean-Baptiste Lespiau and Maria Tsimpoukelli and Nikolai Grigorev and Doug Fritz and Thibault Sottiaux and Mantas Pajarskas and Toby Pohlen and Zhitao Gong and Daniel Toyama and Cyprien de Masson d'Autume and Yujia Li and Tayfun Terzi and Vladimir Mikulik and Igor Babuschkin and Aidan Clark and Diego de Las Casas and Aurelia Guy and Chris Jones and James Bradbury and Matthew Johnson and Blake Hechtman and Laura Weidinger and Iason Gabriel and William Isaac and Ed Lockhart and Simon Osindero and Laura Rimell and Chris Dyer and Oriol Vinyals and Kareem Ayoub and Jeff Stanway and Lorrayne Bennett and Demis Hassabis and Koray Kavukcuoglu and Geoffrey Irving},
      year={2022},
      eprint={2112.11446},
      archivePrefix={arXiv},
      primaryClass={cs.CL},
      url={https://arxiv.org/abs/2112.11446}, 
}

@INPROCEEDINGS{666900,
  author={Broder, A.Z.},
  booktitle={Proceedings. Compression and Complexity of SEQUENCES 1997}, 
  title={On the resemblance and containment of documents}, 
  year={1997},
  volume={},
  number={},
  pages={21-29},
  doi={10.1109/SEQUEN.1997.666900}
}

@misc{eurollm9btech,
      title={EuroLLM-9B: Technical Report}, 
      author={Pedro Henrique Martins and João Alves and Patrick Fernandes and Nuno M. Guerreiro and Ricardo Rei and Amin Farajian and Mateusz Klimaszewski and Duarte M. Alves and José Pombal and Nicolas Boizard and Manuel Faysse and Pierre Colombo and François Yvon and Barry Haddow and José G. C. de Souza and Alexandra Birch and André F. T. Martins},
      year={2025},
      eprint={2506.04079},
      archivePrefix={arXiv},
      primaryClass={cs.CL},
      url={https://arxiv.org/abs/2506.04079}, 
}

@misc{su2025nemotroncctransformingcommoncrawl,
      title={Nemotron-CC: Transforming Common Crawl into a Refined Long-Horizon Pretraining Dataset}, 
      author={Dan Su and Kezhi Kong and Ying Lin and Joseph Jennings and Brandon Norick and Markus Kliegl and Mostofa Patwary and Mohammad Shoeybi and Bryan Catanzaro},
      year={2025},
      eprint={2412.02595},
      archivePrefix={arXiv},
      primaryClass={cs.CL},
      url={https://arxiv.org/abs/2412.02595}, 
}

@article{radford2019language,
  title={Language Models are Unsupervised Multitask Learners},
  author={Radford, Alec and Wu, Jeff and Child, Rewon and Luan, David and Amodei, Dario and Sutskever, Ilya},
  year={2019}
}

@misc{penedo2023refinedwebdatasetfalconllm,
      title={The RefinedWeb Dataset for Falcon LLM: Outperforming Curated Corpora with Web Data, and Web Data Only}, 
      author={Guilherme Penedo and Quentin Malartic and Daniel Hesslow and Ruxandra Cojocaru and Alessandro Cappelli and Hamza Alobeidli and Baptiste Pannier and Ebtesam Almazrouei and Julien Launay},
      year={2023},
      eprint={2306.01116},
      archivePrefix={arXiv},
      primaryClass={cs.CL},
      url={https://arxiv.org/abs/2306.01116}, 
}

@misc{penedo2024finewebdatasetsdecantingweb,
      title={The FineWeb Datasets: Decanting the Web for the Finest Text Data at Scale}, 
      author={Guilherme Penedo and Hynek Kydlíček and Loubna Ben allal and Anton Lozhkov and Margaret Mitchell and Colin Raffel and Leandro Von Werra and Thomas Wolf},
      year={2024},
      eprint={2406.17557},
      archivePrefix={arXiv},
      primaryClass={cs.CL},
      url={https://arxiv.org/abs/2406.17557}, 
}

@inproceedings{xue-etal-2021-mt5,
    title = "m{T}5: A Massively Multilingual Pre-trained Text-to-Text Transformer",
    author = "Xue, Linting  and
      Constant, Noah  and
      Roberts, Adam  and
      Kale, Mihir  and
      Al-Rfou, Rami  and
      Siddhant, Aditya  and
      Barua, Aditya  and
      Raffel, Colin",
    editor = "Toutanova, Kristina  and
      Rumshisky, Anna  and
      Zettlemoyer, Luke  and
      Hakkani-Tur, Dilek  and
      Beltagy, Iz  and
      Bethard, Steven  and
      Cotterell, Ryan  and
      Chakraborty, Tanmoy  and
      Zhou, Yichao",
    booktitle = "Proceedings of the 2021 Conference of the North American Chapter of the Association for Computational Linguistics: Human Language Technologies",
    month = jun,
    year = "2021",
    address = "Online",
    publisher = "Association for Computational Linguistics",
    url = "https://aclanthology.org/2021.naacl-main.41/",
    doi = "10.18653/v1/2021.naacl-main.41",
    pages = "483--498"
}

@inproceedings{abadji-etal-2022-towards,
    title = "Towards a Cleaner Document-Oriented Multilingual Crawled Corpus",
    author = "Abadji, Julien  and
      Ortiz Suarez, Pedro  and
      Romary, Laurent  and
      Sagot, Beno{\^i}t",
    editor = "Calzolari, Nicoletta  and
      B{\'e}chet, Fr{\'e}d{\'e}ric  and
      Blache, Philippe  and
      Choukri, Khalid  and
      Cieri, Christopher  and
      Declerck, Thierry  and
      Goggi, Sara  and
      Isahara, Hitoshi  and
      Maegaard, Bente  and
      Mariani, Joseph  and
      Mazo, H{\'e}l{\`e}ne  and
      Odijk, Jan  and
      Piperidis, Stelios",
    booktitle = "Proceedings of the Thirteenth Language Resources and Evaluation Conference",
    month = jun,
    year = "2022",
    address = "Marseille, France",
    publisher = "European Language Resources Association",
    url = "https://aclanthology.org/2022.lrec-1.463/",
    pages = "4344--4355"
}

@article{JMLR:v21:20-074,
  author  = {Colin Raffel and Noam Shazeer and Adam Roberts and Katherine Lee and Sharan Narang and Michael Matena and Yanqi Zhou and Wei Li and Peter J. Liu},
  title   = {Exploring the Limits of Transfer Learning with a Unified Text-to-Text Transformer},
  journal = {Journal of Machine Learning Research},
  year    = {2020},
  volume  = {21},
  number  = {140},
  pages   = {1--67},
  url     = {http://jmlr.org/papers/v21/20-074.html}
}

@inproceedings{lee-etal-2022-deduplicating,
    title = "Deduplicating Training Data Makes Language Models Better",
    author = "Lee, Katherine  and
      Ippolito, Daphne  and
      Nystrom, Andrew  and
      Zhang, Chiyuan  and
      Eck, Douglas  and
      Callison-Burch, Chris  and
      Carlini, Nicholas",
    editor = "Muresan, Smaranda  and
      Nakov, Preslav  and
      Villavicencio, Aline",
    booktitle = "Proceedings of the 60th Annual Meeting of the Association for Computational Linguistics (Volume 1: Long Papers)",
    month = may,
    year = "2022",
    address = "Dublin, Ireland",
    publisher = "Association for Computational Linguistics",
    url = "https://aclanthology.org/2022.acl-long.577/",
    doi = "10.18653/v1/2022.acl-long.577",
    pages = "8424--8445"
}

@inproceedings{nguyen-etal-2024-culturax,
    title = "{C}ultura{X}: A Cleaned, Enormous, and Multilingual Dataset for Large Language Models in 167 Languages",
    author = "Nguyen, Thuat  and
      Nguyen, Chien Van  and
      Lai, Viet Dac  and
      Man, Hieu  and
      Ngo, Nghia Trung  and
      Dernoncourt, Franck  and
      Rossi, Ryan A.  and
      Nguyen, Thien Huu",
    editor = "Calzolari, Nicoletta  and
      Kan, Min-Yen  and
      Hoste, Veronique  and
      Lenci, Alessandro  and
      Sakti, Sakriani  and
      Xue, Nianwen",
    booktitle = "Proceedings of the 2024 Joint International Conference on Computational Linguistics, Language Resources and Evaluation (LREC-COLING 2024)",
    month = may,
    year = "2024",
    address = "Torino, Italia",
    publisher = "ELRA and ICCL",
    url = "https://aclanthology.org/2024.lrec-main.377/",
    pages = "4226--4237"
}

@inproceedings{henriksson-etal-2025-finerweb,
    title = "{FinerWeb-10BT}: {Refining} Web Data with {LLM}-Based Line-Level Filtering",
    author = "Henriksson, Erik  and
      Tarkka, Otto  and
      Ginter, Filip",
    editor = "Johansson, Richard  and
      Stymne, Sara",
    booktitle = "Proceedings of the Joint 25th Nordic Conference on Computational Linguistics and 11th Baltic Conference on Human Language Technologies (NoDaLiDa/Baltic-HLT 2025)",
    month = mar,
    year = "2025",
    address = "Tallinn, Estonia",
    publisher = "University of Tartu Library",
    url = "https://aclanthology.org/2025.nodalida-1.27/",
    pages = "258--268",
    ISBN = "978-9908-53-109-0"
}

@inproceedings{NEURIPS2022_ce9e92e3,
 author = {Lauren\c{c}on, Hugo and Saulnier, Lucile and Wang, Thomas and Akiki, Christopher and Villanova del Moral, Albert and Le Scao, Teven and Von Werra, Leandro and Mou, Chenghao and Gonz\'{a}lez Ponferrada, Eduardo and Nguyen, Huu and Frohberg, J\"{o}rg and \v{S}a\v{s}ko, Mario and Lhoest, Quentin and McMillan-Major, Angelina and Dupont, Gerard and Biderman, Stella and Rogers, Anna and Ben allal, Loubna and De Toni, Francesco and Pistilli, Giada and Nguyen, Olivier and Nikpoor, Somaieh and Masoud, Maraim and Colombo, Pierre and de la Rosa, Javier and Villegas, Paulo and Thrush, Tristan and Longpre, Shayne and Nagel, Sebastian and Weber, Leon and Mu\~{n}oz, Manuel and Zhu, Jian and Van Strien, Daniel and Alyafeai, Zaid and Almubarak, Khalid and Vu, Minh Chien and Gonzalez-Dios, Itziar and Soroa, Aitor and Lo, Kyle and Dey, Manan and Ortiz Suarez, Pedro and Gokaslan, Aaron and Bose, Shamik and Adelani, David and Phan, Long and Tran, Hieu and Yu, Ian and Pai, Suhas and Chim, Jenny and Lepercq, Violette and Ilic, Suzana and Mitchell, Margaret and Luccioni, Sasha Alexandra and Jernite, Yacine},
 booktitle = {Advances in Neural Information Processing Systems},
 editor = {S. Koyejo and S. Mohamed and A. Agarwal and D. Belgrave and K. Cho and A. Oh},
 pages = {31809--31826},
 publisher = {Curran Associates, Inc.},
 title = {The BigScience ROOTS Corpus: A 1.6TB Composite Multilingual Dataset},
 url = {https://proceedings.neurips.cc/paper_files/paper/2022/file/ce9e92e3de2372a4b93353eb7f3dc0bd-Paper-Datasets_and_Benchmarks.pdf},
 volume = {35},
 year = {2022}
}

@inproceedings{ali-etal-2025-judging,
    title = "Judging Quality Across Languages: A Multilingual Approach to Pretraining Data Filtering with Language Models",
    author = {Ali, Mehdi  and
      Brack, Manuel  and
      L{\"u}bbering, Max  and
      Wendt, Elias  and
      Khan, Abbas Goher  and
      Rutmann, Richard  and
      Jude, Alex  and
      Kraus, Maurice  and
      Weber, Alexander Arno  and
      Stollenwerk, Felix  and
      Kacz{\'e}r, David  and
      Mai, Florian  and
      Flek, Lucie  and
      Sifa, Rafet  and
      Flores-Herr, Nicolas  and
      Koehler, Joachim  and
      Schramowski, Patrick  and
      Fromm, Michael  and
      Kersting, Kristian},
    editor = "Christodoulopoulos, Christos  and
      Chakraborty, Tanmoy  and
      Rose, Carolyn  and
      Peng, Violet",
    booktitle = "Proceedings of the 2025 Conference on Empirical Methods in Natural Language Processing",
    month = nov,
    year = "2025",
    address = "Suzhou, China",
    publisher = "Association for Computational Linguistics",
    url = "https://aclanthology.org/2025.emnlp-main.449/",
    doi = "10.18653/v1/2025.emnlp-main.449",
    pages = "8859--8898",
    ISBN = "979-8-89176-332-6"
}

@misc{datologyai2026uberwebinsightsmultilingualcuration,
      title={\"UberWeb: Insights from Multilingual Curation for a 20-Trillion-Token Dataset}, 
      author={DatologyAI and : and Aldo Gael Carranza and Kaleigh Mentzer and Ricardo Pio Monti and Alex Fang and Alvin Deng and Amro Abbas and Anshuman Suri and Brett Larsen and Cody Blakeney and Darren Teh and David Schwab and Diego Kiner and Fan Pan and Haakon Mongstad and Haoli Yin and Jack Urbanek and Jason Lee and Jason Telanoff and Josh Wills and Luke Merrick and Maximilian Böther and Parth Doshi and Paul Burstein and Pratyush Maini and Rishabh Adiga and Siddharth Joshi and Spandan Das and Tony Jiang and Vineeth Dorna and Zhengping Wang and Bogdan Gaza and Ari Morcos and Matthew Leavitt},
      year={2026},
      eprint={2602.15210},
      archivePrefix={arXiv},
      primaryClass={cs.LG},
      url={https://arxiv.org/abs/2602.15210}, 
}

@inproceedings{burns-etal-2026-aleph,
    title = "Aleph-Alpha-{G}erman{W}eb: Improving {G}erman-language {LLM} pre-training with model-based data curation and synthetic data generation",
    author = {Burns, Thomas F  and
      Parcalabescu, Letitia  and
      Waeldchen, Stephan  and
      Barlow, Michael  and
      Ziegltrum, Gregor  and
      Stampa, Volker  and
      Harren, Bastian  and
      Deiseroth, Bj{\"o}rn},
    editor = "Demberg, Vera  and
      Inui, Kentaro  and
      Marquez, Llu{\'i}s",
    booktitle = "Proceedings of the 19th Conference of the {E}uropean Chapter of the {A}ssociation for {C}omputational {L}inguistics (Volume 1: Long Papers)",
    month = mar,
    year = "2026",
    address = "Rabat, Morocco",
    publisher = "Association for Computational Linguistics",
    url = "https://aclanthology.org/2026.eacl-long.58/",
    doi = "10.18653/v1/2026.eacl-long.58",
    pages = "1267--1283",
    ISBN = "979-8-89176-380-7"
}

@inproceedings{ICLR2025_cde43c5d,
 author = {Norlund, Tobias and Isbister, Tim and Cuba Gyllensten, Amaru and dos Santos, Paul and Petrelli, Danila and Ekgren, Ariel and Sahlgren, Magnus},
 booktitle = {International Conference on Learning Representations},
 editor = {Y. Yue and A. Garg and N. Peng and F. Sha and R. Yu},
 pages = {82775--82798},
 title = {SWEb: A Large Web Dataset for the Scandinavian Languages},
 url = {https://proceedings.iclr.cc/paper_files/paper/2025/file/cde43c5d698076ae17f3838a80f5a1bc-Paper-Conference.pdf},
 volume = {2025},
 year = {2025}
}

@misc{almeida2025buildinghighqualitydatasetsportuguese,
      title={Building High-Quality Datasets for Portuguese LLMs: From Common Crawl Snapshots to Industrial-Grade Corpora}, 
      author={Thales Sales Almeida and Rodrigo Nogueira and Helio Pedrini},
      year={2025},
      eprint={2509.08824},
      archivePrefix={arXiv},
      primaryClass={cs.CL},
      url={https://arxiv.org/abs/2509.08824}, 
}

@inproceedings{simplicio-etal-2026-amalia,
    title = "{AMALIA}: A Fully Open Large Language Model for {E}uropean {P}ortuguese",
    author = "Simpl{\'i}cio, Afonso  and
      Vinagre, Gon{\c{c}}alo  and
      Ramos, Miguel Moura  and
      Tavares, Diogo  and
      Ferreira, Rafael  and
      Attanasio, Giuseppe  and
      Alves, Duarte M.  and
      Calvo, In{\^e}s  and
      Vieira, In{\^e}s  and
      Guerra, Rui  and
      Furtado, James  and
      Canaverde, Beatriz  and
      Paulo, Iago  and
      Ramos, Vasco  and
      Gl{\'o}ria-Silva, Diogo  and
      Faria, Miguel  and
      Treviso, Marcos  and
      Gomes, Daniel  and
      Gomes, Pedro  and
      Semedo, David  and
      Martins, Andr{\'e}  and
      Magalh{\~a}es, Jo{\~a}o",
    editor = "Souza, Marlo  and
      de-Dios-Flores, Iria  and
      Santos, Diana  and
      Freitas, Larissa  and
      Souza, Jackson Wilke da Cruz  and
      Ribeiro, Eug{\'e}nio",
    booktitle = "Proceedings of the 17th International Conference on Computational Processing of {P}ortuguese ({PROPOR} 2026) - Vol. 1",
    month = apr,
    year = "2026",
    address = "Salvador, Brazil",
    publisher = "Association for Computational Linguistics",
    url = "https://aclanthology.org/2026.propor-1.38/",
    pages = "380--391",
    ISBN = "979-8-89176-387-6"
}

@inproceedings{vieira-etal-2026-alba,
    title = "{ALBA}: A {E}uropean {P}ortuguese Benchmark for Evaluating Language and Linguistic Dimensions in Generative {LLM}s",
    author = "Vieira, In{\^e}s  and
      Calvo, In{\^e}s  and
      Paulo, Iago  and
      Furtado, James  and
      Ferreira, Rafael  and
      Tavares, Diogo  and
      Gl{\'o}ria-Silva, Diogo  and
      Semedo, David  and
      Magalh{\~a}es, Jo{\~a}o",
    editor = "Souza, Marlo  and
      de-Dios-Flores, Iria  and
      Santos, Diana  and
      Freitas, Larissa  and
      Souza, Jackson Wilke da Cruz  and
      Ribeiro, Eug{\'e}nio",
    booktitle = "Proceedings of the 17th International Conference on Computational Processing of {P}ortuguese ({PROPOR} 2026) - Vol. 1",
    month = apr,
    year = "2026",
    address = "Salvador, Brazil",
    publisher = "Association for Computational Linguistics",
    url = "https://aclanthology.org/2026.propor-1.69/",
    pages = "697--707",
    ISBN = "979-8-89176-387-6"
}

@misc{abbas2023semdedupdataefficientlearningwebscale,
      title={SemDeDup: Data-efficient learning at web-scale through semantic deduplication}, 
      author={Amro Abbas and Kushal Tirumala and Dániel Simig and Surya Ganguli and Ari S. Morcos},
      year={2023},
      eprint={2303.09540},
      archivePrefix={arXiv},
      primaryClass={cs.LG},
      url={https://arxiv.org/abs/2303.09540}, 
}

@inproceedings{ferreira-etal-2026-p3b3,
    title = "{P}3{B}3: A Multi-Turn Conversational Benchmark for Measuring {E}uropean and {B}razilian {P}ortuguese Variety Bias in {LLM}s",
    author = "Ferreira, Rafael  and
      Vieira, In{\^e}s  and
      Calvo, In{\^e}s  and
      Furtado, James  and
      Paulo, Iago  and
      Gl{\'o}ria-Silva, Diogo  and
      Tavares, Diogo  and
      Semedo, David  and
      Magalhaes, Joao",
    editor = "Huang, Kaiyu  and
      Mo, Fengran  and
      Chen, Pinzhen  and
      Jiang, Meng",
    booktitle = "Proceedings of the 1st Workshop on Multilinguality in the Era of Large Language Models ({M}e{LLM} 2026)",
    month = jul,
    year = "2026",
    address = "San Diego, United States",
    publisher = "Association for Computational Linguistics",
    url = "https://aclanthology.org/2026.mellm-1.23/",
    doi = "10.18653/v1/2026.mellm-1.23",
    pages = "240--248",
    ISBN = "979-8-89176-430-9"
}

@article{grattafiori2024llama,
  title={The Llama 3 Herd of Models},
  author={Dubey, Abhimanyu and Jauhri, Abhinav and Pandey, Abhinav and Kadian, Abhishek and Al-Dahle, Ahmad and Letman, Aiesha and Mathur, Akhil and Schelten, Alan and Yang, Amy and Fan, Angela and others},
  journal={arXiv preprint arXiv:2407.21783},
  url={https://arxiv.org/abs/2407.21783},
  year={2024}
}

@article{OLMo20242O2,
  title={2 OLMo 2 Furious},
  author={Team OLMo and Pete Walsh and Luca Soldaini and Dirk Groeneveld and Kyle Lo and Shane Arora and Akshita Bhagia and Yuling Gu and Shengyi Huang and Matt Jordan and Nathan Lambert and Dustin Schwenk and Oyvind Tafjord and Taira Anderson and David Atkinson and Faeze Brahman and Christopher Clark and Pradeep Dasigi and Nouha Dziri and Michal Guerquin and Hamish Ivison and Pang Wei Koh and Jiacheng Liu and Saumya Malik and William Merrill and Lester James Validad Miranda and Jacob Daniel Morrison and Tyler C. Murray and Crystal Nam and Valentina Pyatkin and Aman Rangapur and Michael Schmitz and Sam Skjonsberg and David Wadden and Christopher Wilhelm and Michael Wilson and Luke S. Zettlemoyer and Ali Farhadi and Noah A. Smith and Hanna Hajishirzi},
  journal={ArXiv},
  year={2024},
  volume={abs/2501.00656},
  url={https://api.semanticscholar.org/CorpusID:275213098}
}

@inproceedings{tavares-etal-2026-pheb,
  title = {PHEB: An European Portuguese High School-Level LLM Benchmark},
  author = {Tavares, Diogo C. and Ferreira, Rafael and Simplício, Afonso and Vinagre, Gonçalo and Condez, Ana Carolina and Calvo, Inês and Vieira, Inês and Semedo, David and Magalhaes, Joao},
  booktitle = {Proceedings of the Fifteenth Language Resources and Evaluation Conference (LREC 2026)},
  month = {May},
  year = {2026},
  pages = {4673--4683},
  address = {Palma, Mallorca, Spain},
  publisher = {European Language Resources Association (ELRA)},
  editor = {Piperidis, Stelios and Bel, Núria and van den Heuvel, Henk and Ide, Nancy and Krek, Simon and Toral, Antonio},
  doi = {10.63317/2o3fvueefvwj}
}

@inproceedings{verbatim-memo,
  author       = {Jing Huang and
                  Diyi Yang and
                  Christopher Potts},
  editor       = {Yaser Al{-}Onaizan and
                  Mohit Bansal and
                  Yun{-}Nung Chen},
  title        = {Demystifying Verbatim Memorization in Large Language Models},
  booktitle    = {Proceedings of the 2024 Conference on Empirical Methods in Natural
                  Language Processing, {EMNLP} 2024, Miami, FL, USA, November 12-16,
                  2024},
  pages        = {10711--10732},
  publisher    = {Association for Computational Linguistics},
  year         = {2024},
  url          = {https://doi.org/10.18653/v1/2024.emnlp-main.598},
  doi          = {10.18653/V1/2024.EMNLP-MAIN.598},
  bibsource    = {dblp computer science bibliography, https://dblp.org}
}

@inproceedings{baucells-etal-2025-iberobench,
    title = "{I}bero{B}ench: A Benchmark for {LLM} Evaluation in {I}berian Languages",
    author = "Baucells, Irene  and
      Aula-Blasco, Javier  and
      de-Dios-Flores, Iria  and
      Paniagua Su{\'a}rez, Silvia  and
      Perez, Naiara  and
      Salles, Anna  and
      Sotelo Docio, Susana  and
      Falc{\~a}o, J{\'u}lia  and
      Saiz, Jose Javier  and
      Sepulveda Torres, Robiert  and
      Barnes, Jeremy  and
      Gamallo, Pablo  and
      Gonzalez-Agirre, Aitor  and
      Rigau, German  and
      Villegas, Marta",
    editor = "Rambow, Owen  and
      Wanner, Leo  and
      Apidianaki, Marianna  and
      Al-Khalifa, Hend  and
      Eugenio, Barbara Di  and
      Schockaert, Steven",
    booktitle = "Proceedings of the 31st International Conference on Computational Linguistics",
    month = jan,
    year = "2025",
    address = "Abu Dhabi, UAE",
    publisher = "Association for Computational Linguistics",
    url = "https://aclanthology.org/2025.coling-main.699/",
    pages = "10491--10519"
}

@inproceedings{preda-etal-2024-across,
    title = "Across the Atlantic: Distinguishing Between {E}uropean and {B}razilian {P}ortuguese Dialects",
    author = "Preda, David  and
      Os{\'o}rio, Tom{\'a}s  and
      Cardoso, Henrique Lopes",
    editor = "Gamallo, Pablo  and
      Claro, Daniela  and
      Teixeira, Ant{\'o}nio  and
      Real, Livy  and
      Garcia, Marcos  and
      Oliveira, Hugo Gon{\c{c}}alo  and
      Amaro, Raquel",
    booktitle = "Proceedings of the 16th International Conference on Computational Processing of Portuguese - Vol. 1",
    month = mar,
    year = "2024",
    address = "Santiago de Compostela, Galicia/Spain",
    publisher = "Association for Computational Lingustics",
    url = "https://aclanthology.org/2024.propor-1.36/",
    pages = "353--363"
}

@misc{euptvid2026,
  author = {Duarte O. Carmo},
  title = {fasttext-euptvid: Fast Portuguese Variety Identification},
  year = {2026},
  url = {https://huggingface.co/duarteocarmo/fasttext-euptvid}
}

\appendix

\clearpage
\section{Collection Selection Details}
\label{app:col-selection}
As explained in Section~\ref{sec:data-collection}, Arquivo.pt comprises several collection types, with varying contents, focuses, and crawling frequency. This variety required us to examine each collection type and define a tailored processing procedure for each.

The FAWP collections contain websites with frequently updated content, such as news media, public and governmental pages, which were crawled every day and saved as quarterly collections. Given the highly variable content, this was the first collection type we processed, using almost every quarterly collection between the 2010-2016 and 2020-2024 time periods.

In order to vary the type of content, we also started processing AWP collections, which are complete crawls of the Portuguese Web, mainly from the .pt top level domain. They contain mainly blogs and other general content that was crawled and saved quarterly. Given the larger volume and higher chance of duplicate content, we selected roughly one collection of similar size per year in the 2008-2024 time period.

Lastly, we also selected to process an assortment of smaller collections, including high-quality, manually curated pages, donated pages, pages crawled during special events, and pages archived by Arquivo.pt users with their ArchivePageNow\footnote{\url{https://arquivo.pt/archivepagenow}} service. These pages, collected during the 1997-2024 time period, were used to further improve the variety and quality of the extracted data.

\section{Dropped Samples per Stage}
\begin{figure*}
    \centering
    \includegraphics[
        width=0.8\linewidth,
        trim={0.2cm 0.2cm 0.2cm 0.2cm},
        clip
    ]{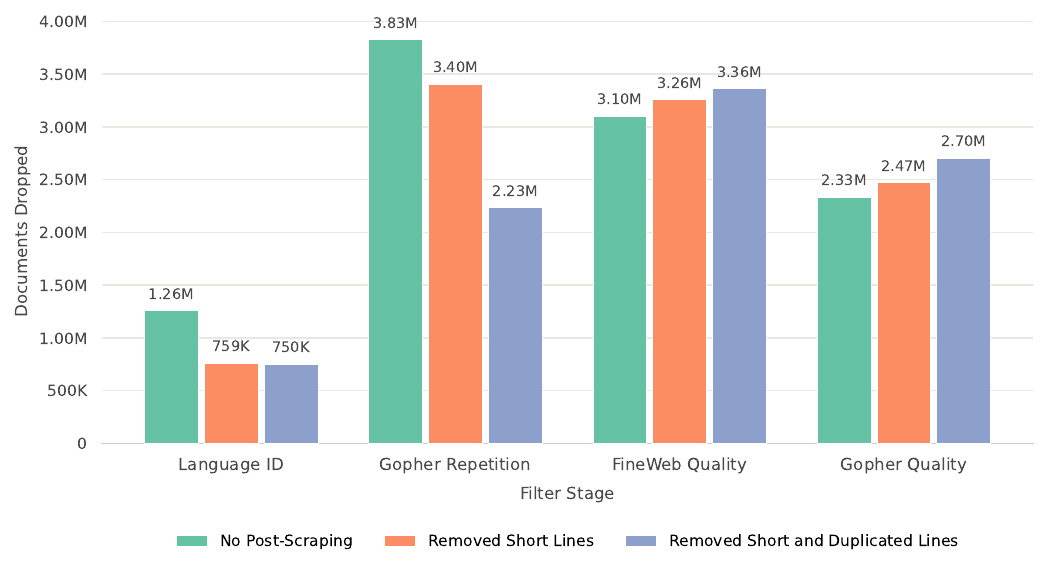}
    \caption{Comparison of the dropped documents with the different post-scraping approaches, per filtering stage. }
    \label{fig:dropped-per-stage}
\end{figure*}

As explained in Section~\ref{sec:post-scraping_experiments}, the documents that are preserved in the early stages are still subject to the subsequent heuristic filters, and many are removed by them, demonstrating the effectiveness of these post-scraping steps. Figure~\ref{fig:dropped-per-stage} shows this evolution per stage.

\section{Rescued Samples by the Post-Scraping Stage}
\label{app:rescued_samples}

Table~\ref{tab:post-scraping-examples} shows examples of documents rescued by the post-scraping step described in Section~\ref{sec:post-scraping_experiments}. Each example shows a document in its raw scraped form and the resulting document after subjected to the post-scraping step. This table illustrates the range of cases the post-scraping step addresses, each of which would otherwise cause the document to be discarded by the downstream heuristic filters.

\begin{table*}[t]
\centering
\small
\begin{tabular}{p{0.48\textwidth}p{0.48\textwidth}}
\toprule
\textbf{Before Post-Scraping} & \textbf{After Post-Scraping} \\
\midrule
\multicolumn{2}{c}{\textit{Duplicated paragraph}} \\
\midrule
\begin{minipage}[t]{\linewidth}
Royal Blood ou a energia do rock\textbackslash nNo último concerto desta digressão, a banda de rock britânica encheu quinta-feira o Coliseu dos Recreios, em Lisboa.\textbackslash nRoyal Blood ou a energia do rock\textbackslash nNo último concerto desta digressão, a banda de rock britânica encheu quinta-feira o Coliseu dos Recreios, em Lisboa.
\end{minipage} &
\begin{minipage}[t]{\linewidth}
Royal Blood ou a energia do rock\textbackslash nNo último concerto desta digressão, a banda de rock britânica encheu quinta-feira o Coliseu dos Recreios, em Lisboa.
\end{minipage} \\
\addlinespace
\midrule
\multicolumn{2}{c}{\textit{Repeated boilerplate label}} \\
\midrule
\begin{minipage}[t]{\linewidth}
Vida e Carreira\textbackslash nVida e Carreira\textbackslash nVida e Carreira\textbackslash nVida e Carreira\textbackslash nJuntámos tudo o que gosta num só site! Mais simples, mais confortável e disponível também no seu telemóvel ou tablet. Esperemos que goste e que continue a acompanhar-nos diariamente.
\end{minipage} &
\begin{minipage}[t]{\linewidth}
Vida e Carreira\textbackslash nJuntámos tudo o que gosta num só site! Mais simples, mais confortável e disponível também no seu telemóvel ou tablet. Esperemos que goste e que continue a acompanhar-nos diariamente.
\end{minipage} \\
\addlinespace
\midrule
\multicolumn{2}{c}{\textit{Pagination header and footer}} \\
\midrule
\begin{minipage}[t]{\linewidth}
Notícias\textbackslash nLeiria: GNR afasta hipótese de crime na morte de casal\textbackslash nEdição de:\textbackslash nAs autoridades afastam a hipótese de ter ocorrido um crime na morte de um casal que foi encontrado dentro de uma viatura na passada sexta-feira à tarde num pinhal do Coimbrão, concelho de Leiria.\textbackslash nPáginas\textbackslash n- « primeira\textbackslash n- ‹ anterior\textbackslash n- 1\textbackslash n- 2\textbackslash n- 3\textbackslash n- 4\textbackslash n- 5\textbackslash n- 6\textbackslash n- 7
\end{minipage} &
\begin{minipage}[t]{\linewidth}
Leiria: GNR afasta hipótese de crime na morte de casal\textbackslash nAs autoridades afastam a hipótese de ter ocorrido um crime na morte de um casal que foi encontrado dentro de uma viatura na passada sexta-feira à tarde num pinhal do Coimbrão, concelho de Leiria.
\end{minipage} \\
\addlinespace
\midrule
\multicolumn{2}{c}{\textit{Low alphabetic ratio}} \\
\midrule
\begin{minipage}[t]{\linewidth}
Ribeiro Cristóvão\textbackslash n19-03-2015 07:24:37\textbackslash nO crivo da Liga dos Campeões apenas deixou passar para os quartos-de-final da competição o Futebol Clube do Porto, Real Madrid, Barcelona, Atlético de Madrid, Bayern de Munique, Juventus, Mónaco e Paris Saint-Germain.
\end{minipage} &
\begin{minipage}[t]{\linewidth}
Ribeiro Cristóvão\textbackslash nO crivo da Liga dos Campeões apenas deixou passar para os quartos-de-final da competição o Futebol Clube do Porto, Real Madrid, Barcelona, Atlético de Madrid, Bayern de Munique, Juventus, Mónaco e Paris Saint-Germain.
\end{minipage} \\
\addlinespace
\midrule
\multicolumn{2}{c}{\textit{Structural/navigation markers}} \\
\midrule
\begin{minipage}[t]{\linewidth}
|\textbackslash n2015-06-12\textbackslash nAssociações de Pais defendem que alunos deviam ter apenas um mês de férias, concorda?\textbackslash n|\textbackslash n|\textbackslash n|\textbackslash n|\textbackslash n2015-06-09\textbackslash nRui Vitória é a melhor opção para o Benfica?\textbackslash n|\textbackslash n|\textbackslash n|\textbackslash n|\textbackslash n2015-06-04\textbackslash nQuem ficou a ganhar com a mudança de Jesus para Alvalade?\textbackslash n|\textbackslash n|\textbackslash n|\textbackslash n|\textbackslash n2015-05-31\textbackslash nO Sporting é um justo vencedor da Taça de Portugal?\textbackslash n|\textbackslash n|\textbackslash n|
\end{minipage} &
\begin{minipage}[t]{\linewidth}
Associações de Pais defendem que alunos deviam ter apenas um mês de férias, concorda?\textbackslash nRui Vitória é a melhor opção para o Benfica?\textbackslash nQuem ficou a ganhar com a mudança de Jesus para Alvalade?\textbackslash nO Sporting é um justo vencedor da Taça de Portugal?
\end{minipage} \\
\bottomrule
\end{tabular}
\caption{Examples of documents rescued by the post-scraping step, comparing their raw form to the result after post-scraping. Each example is labeled with the issue that would otherwise have caused the document to be discarded by the heuristic filters.}
\label{tab:post-scraping-examples}
\end{table*}

\section{Removed Samples}
\label{app:removed-samples}

Table~\ref{tab:discarded-samples} shows an example of a discarded sample for each filter of the filter stages.

\definecolor{Alto}{rgb}{0.819,0.819,0.819}
\definecolor{Nobel}{rgb}{0.713,0.713,0.713}
\begin{table*}
\centering
\footnotesize
\begin{tblr}{
  colspec = {
    Q[c, m, wd=0.117\linewidth] | 
    Q[l, m, wd=0.1\linewidth] 
    X[l, m]
  },
  row{1} = {font=\bfseries},
  row{3} = {t},
  row{4} = {t},
  row{5} = {t},
  row{6} = {t},
  row{7} = {t},
  row{8} = {t},
  row{9} = {t},
  row{10} = {t},
  row{12} = {t},
  row{14} = {t},
  row{15} = {t},
  row{16} = {t},
  row{17} = {t},
  row{18} = {t},
  cell{1}{1} = {c},
  cell{2}{1} = {r=9}{c},
  cell{2}{2} = {t},
  cell{2}{3} = {t},
  cell{11}{1} = {r=2}{c},
  cell{11}{2} = {t},
  cell{11}{3} = {t},
  cell{13}{1} = {r=6}{c},
  cell{13}{2} = {t},
  cell{13}{3} = {t},
  vline{2} = {1-18}{},
  hline{2,19} = {-}{},
  hline{3} = {2-3}{Nobel},
  hline{4-10,12,14-18} = {2-3}{Alto},
  hline{11,13} = {-}{Nobel},
}
Filter Stage & Filtering Reason & Discarded Text\\
Gopher Repetition & Top 2-gram & Rever episódios rtp.pt/beirais\textbackslash{}nRever episódios rtp.pt/aguademar\textbackslash{}nRever Episódios\textbackslash{}nRever episódios rtp.pt/osfilhosdorock\textbackslash{}nRever episódios\\
 & Top 3-gram & Resultados e Classificações\textbackslash{}nUEFA Champions League 2014/2015\textbackslash{}nC. Amarelos \textbar{} C. Vermelhos \textbar{}\textbackslash{}nMelhor Ataque \textbar{}\textbackslash{}nMelhor Defesa \textbar{}\textbackslash{}nESTATÍSTICAS DA UEFA Champions League\textbackslash{}nClassificação\\
 & Top 4-gram & A Microsoft quer estrear no Windows 10 uma loja única para comercializar vídeos, apps e músicas.\textbackslash{}nLer mais: Windows 10 vai ter uma loja única para vídeos, músicas e apps\\
 & Duplicated\newline 5-grams & Os GNR voltam ao palco do Coliseu do Porto.\textbackslash{}nOs GNR voltam ao palco do Coliseu dos Recreios.\textbackslash{}nOs Apocalyptica dão um concerto no Coliseu do Porto para apresentar o novo álbum.\textbackslash{}nOs Apocalyptica dão um concerto no Coliseu dos Recreios para apresentar o novo álbum.\\
 & Duplicated\newline 6-grams & Diana Chaves na capa da 'Women's Health'\textbackslash{}nBom dia!!! Já viram quem está na capa da Women's Health Portugal deste mês??\textbackslash{}n"Bom dia!!! Já viram quem está na capa da Women's Helath Portugal deste mês??", escreveu a atriz e apresentadora Diana Chaves na legenda da foto que publicou no Facebook.\\
 & Duplicated\newline 7-grams & Trailer - Mudar de Vida (versão digital restaurada)\textbackslash{}nTrailer oficial de Mudar de Vida (versão digital restaurada) Com Geraldo Del Rey, Isabel Ruth e María Barroso. Realização de Paulo Rocha.\\
 & Duplicated\newline 8-grams & “Educação para uma Cultura de Segurança”\textbackslash{}nRealizou-se o Colóquio intitulado “Educação para uma Cultura de Segurança”, evento promovido pela Secretaria Regional de Educação e Recursos Humanos, em parceria com a Associação Insular de Geografia. Fotos: FM\\
 & Duplicated\newline 9-grams & Mariana Monteiro aventureira\textbackslash{}nA tirar partido da Natureza , sinto-me como uma criança\textbackslash{}nMariana Monteiro publicou uma fotografia no Facebook, onde aparece em modo desportivo a descontrair na natureza. "A tirar partido da Natureza , sinto-me como uma criança", escreveu na legenda da imagem.\\
 & Duplicated\newline 10-grams & Iva Lamarão e André Santos mais apaixonados que nunca\textbackslash{}nO mundo é pequeno para o nosso amor. Contigo. Amo te\textbackslash{}n"O mundo é pequeno para o nosso amor. Contigo. Amo-te", escreveu Iva Lamarão na legenda da imagem, onde aparece com o namorado, o futebolista André Santos.\\
FineWeb Quality & Line punctuation ratio & Descarregue a aplicação:\textbackslash{}npublicidade\\
 & List ratio & Que Música Tocou?\textbackslash{}nOS AZEITONAS\textbackslash{}nNOS DESENHOS ANIMADOS\textbackslash{}nWHO'S GONNA RIDE YOUR WILD HORSES\textbackslash{}nCHRISTINA PERRI\textbackslash{}nJAR OF HEARTS\textbackslash{}nSegunda-Feira\textbackslash{}nTerça-Feira\textbackslash{}nQuarta-Feira\textbackslash{}nQuinta-Feira\textbackslash{}nSexta-Feira\\
Gopher Quality & Below alpha threshold & A iniciativa acontece entre hoje e quarta-feira, de 11 a 13 de Maio, e alarga-se a todo o país. Os organizadores esperam atrair, no mínimo, 200 mil espectadores.\textbackslash{}nLer mais: Festa do Cinema com bilhetes a 2,5 euros até quarta-feira\\
 & Short document & Ronaldo pode assumir a liderança dos melhores marcadores\textbackslash{}nA redireccionar...\\
  & Long document & {"O desnorte está a instalar-se nas hostes governamentais e para-governamentais.\textbackslash{}n\textit{etc..}"\\(This discarded document had 237k tokens)}\\
 & Too many end ellipsis & Adriano Lucas (1925-2011)\textbackslash{}nAdriano Callé Lucas\textbackslash{}nHá 25 anos a informar...\textbackslash{}nA câmara do Bombarral aprovou ontem um Plano de Emergência Social para ajudar carenciados a pagar despesas básicas, como medicamentos, água e...\textbackslash{}nNão existem notícias em destaque.\\
 & Too many bullets & - As opiniões de Alexandre Mestre (Secretário de Estadodo Desporto),Fernando Gomes (Presidente da FPF) Vitor Baía, Rui Costa, Beto e Barroso.\textbackslash{}n- O olhar feminino sobre os quatro "grandes".\textbackslash{}n- Os gostos extrafutebol dos craques da bola.\textbackslash{}n- Para que servem as equipas B dos maiores clubes?\textbackslash{}n- Por que razão arrancam especificamente este ano?\\
 & Not enough stop words & Diário de Notícias\textbackslash{}nNecrologia\textbackslash{}nPARTICIPAÇÃO\textbackslash{}nMaria Ângela Jardim Felgueira\textbackslash{}nSeus irmãos: Maria Inês Jardim Felgueira Soares, seu marido,...
\end{tblr}
\caption{Examples of discarded documents for each filter. This shows the importance of a set of rigorous filtering stages, as these documents would undermine our final corpus.}
\label{tab:discarded-samples}
\end{table*}

\section{Memorization Example}

Table~\ref{tab:mem-examples} show examples where AMALIA memorized more then usual, as these were exceptions of our experiments.

\definecolor{matchgreen}{rgb}{0.0,0.55,0.0}
\definecolor{failred}{rgb}{0.80,0.0,0.0}
\definecolor{truthblue}{rgb}{0.0,0.30,0.80}
\begin{table*}[t]
\centering
\footnotesize
\renewcommand{\arraystretch}{1.2}
\begin{tabularx}{\linewidth}{@{}l >{\raggedright\arraybackslash}X@{}}
\toprule
\multicolumn{2}{@{}l}{\textbf{Prompt} (\emph{italic}); continuation: \textcolor{matchgreen}{verbatim match} / \textcolor{failred}{divergence} / \textcolor{truthblue}{ground truth}.} \\
\midrule
\multicolumn{2}{@{}l}{\textit{Example 1}} \\
\addlinespace[2pt]
\textsc{AMALIA} \emph{(26 w)} & \textit{Se tiver com níveis de tensão arterial elevados, é importante que saiba que isso pode ter [\dots] nas artérias e na circulação. O nível de tensão arterial é considerado elevado quando os} \\
 & \textcolor{matchgreen}{valores de tensão arterial sistólica são iguais ou superiores a 140 mmHg e os valores de tensão arterial diastólica são iguais ou superiores a 90 mmHg. {\color{gray}\textbackslash n}}\textcolor{failred}{A tensão arterial elevada é uma doença silenciosa, uma vez que não apresenta sintomas. Por isso, é importante que faça exames}[\dots] \\
 & \textcolor{truthblue}{Porque é que a tensão arterial é tão importante? {\color{gray}\textbackslash n} É importante que conheça os seus níveis de tensão arterial. A boa notícia é que é}[\dots] \\
\midrule
\multicolumn{2}{@{}l}{\textit{Example 2}} \\
\addlinespace[2pt]
\textsc{AMALIA} \emph{(18 w)} & \textit{Pode consultar Legislação e Outros Documentos de três formas distintas: acedendo a [\dots] Quando se deparar com este símbolo, na secção de “Documentos Legais” ou "Publicações",} \\
 & \textcolor{matchgreen}{saiba que a legislação ou outro tipo de documento estão disponíveis para download, em formato PDF ou DOC. {\color{gray}\textbackslash n}}\textcolor{failred}{Para aceder à legislação ou outros documentos através da secção “Legislação e Regulamentação” deve pressionar em “Legislação e}[\dots] \\
 & \textcolor{truthblue}{Quer na área das “Eu Cidadão e…” quer das “Áreas de Interesse”, pode aceder à legislação relativa ao tema que está a consultar}[\dots] \\
\midrule
\multicolumn{2}{@{}l}{\textit{Example 3}} \\
\addlinespace[2pt]
\textsc{AMALIA} \emph{(12 w)} & \textit{Título: Bocage – a vida apaixonada de um genial libertino {\color{gray}\textbackslash n} Autor: Luís Rosa {\color{gray}\textbackslash n} Editora: [\dots] (passageiros e carga) têm origem e destino, predominantemente, na zona geográfica em que} \\
 & \textcolor{matchgreen}{se encontram localizados; que possuem uma bacia de atracção de tráfego aéreo }\textcolor{failred}{mais extensa; com ligações aéreas regulares para um ou dois aeroportos de tipo superior, não mais – todos eles localizados, por}[\dots] \\
 & \textcolor{truthblue}{já bastante ampla e profunda, servida principalmente por outros meios de transporte; com ligações aéreas regulares a uma}[\dots] \\
\midrule
\multicolumn{2}{@{}l}{\textit{Example 4}} \\
\addlinespace[2pt]
\textsc{AMALIA} \emph{(11 w)} & \textit{«A escola, que tão mal ensina a escrever, não ensina, de todo, a falar.» Quem o afirma é [\dots] e quando a falam.» {\color{gray}\textbackslash n} Lembra, contudo, o que para ele é uma «obsessão»: «(...) A escola,} \\
 & \textcolor{matchgreen}{que tão mal ensina a escrever, não ensina, de todo, a falar.}\textcolor{failred}{» {\color{gray}\textbackslash n} «(...) A língua é um organismo vivo, e como tal, sujeito a mutações, a transformações, a adaptações, a reacções, a reacções}[\dots] \\
 & \textcolor{truthblue}{(...) A aprendizagem elementar da fala e o desenvolvimento da língua estão entregues às famílias, ao meio técnico e cultural em}[\dots] \\
\midrule
\multicolumn{2}{@{}l}{\textit{Example 5}} \\
\addlinespace[2pt]
\textsc{AMALIA} \emph{(10 w)} & \textit{Segunda-feira, 18 de Junho de 2007 {\color{gray}\textbackslash n} É como um mealheiro de sentimentos e de estados de [\dots] que amor são esses, quando uma pessoa segue para a esquerda e a outra para a direita? Que} \\
 & \textcolor{matchgreen}{amizade ou que amor são esses, quando uma pessoa se }\textcolor{failred}{afasta e a outra se aproxima? Que amizade ou que amor são esses, quando uma pessoa se afasta e a outra se afasta? Que amizade ou}[\dots] \\
 & \textcolor{truthblue}{limita a seguir os passos da outra? A amizade e o amor só fazem sentido quando as pessoas caminham no sentido da convergência. Só}[\dots] \\
\bottomrule
\end{tabularx}\caption{The five documents AMALIA reproduces most verbatim, ranked by the number of consecutive matching words. For each, \textcolor{matchgreen}{green} marks the words AMALIA reproduces exactly from the held-out continuation, \textcolor{failred}{red} marks where it diverges, and \textcolor{truthblue}{blue} is the ground-truth continuation. Even in the worst case the model reproduces only 26 consecutive words.}
\label{tab:mem-examples}
\end{table*}

\end{document}